\documentclass[10pt,twocolumn,letterpaper]{article}

\usepackage{iccv}
\usepackage{times}
\usepackage{epsfig}
\usepackage{graphicx}
\usepackage{amsmath}
\usepackage{amssymb}

\usepackage{subcaption}
\usepackage{longtable}
\usepackage{float}
\usepackage[flushleft]{threeparttable}
\usepackage{enumitem}
\usepackage{algorithm}
\usepackage{algorithmic}
\usepackage{listings}

\usepackage{amsmath,amsfonts, mathtools, bm, bbm, nicefrac}

\newcommand{\E}{\mathbb{E}}

\def\vx{{\bm{x}}}
\def\vy{{\bm{y}}}
\def\vz{{\bm{z}}}

\DeclareMathAlphabet{\mathsfit}{\encodingdefault}{\sfdefault}{m}{sl}
\SetMathAlphabet{\mathsfit}{bold}{\encodingdefault}{\sfdefault}{bx}{n}

\usepackage{booktabs}
\usepackage{hyperref}
\usepackage{comment}
\usepackage{amsmath,amssymb} 
\usepackage{color}
\usepackage{booktabs}

\usepackage{bbm}
\usepackage{multicol}
\usepackage{makecell}
\usepackage{xcolor}
\usepackage{xspace}
\usepackage{float}

\usepackage[accsupp]{axessibility}  
\usepackage{graphicx}
\usepackage{animate}
\usepackage[capitalize]{cleveref}
\crefname{section}{Sec.}{Secs.}
\Crefname{section}{Section}{Sections}
\Crefname{table}{Table}{Tables}
\crefname{table}{Tab.}{Tabs.}

\newcommand{\myPara}[1]{\vspace{6pt}\noindent\textbf{#1}}

\newcommand{\tit}[1]{\textit{#1}}

\iccvfinalcopy 

\ificcvfinal\fi

\begin{document}

\title{ MagicVideo: Efficient Video Generation With Latent Diffusion Models}

\author{
  Daquan Zhou$^*$ \quad Weimin Wang\thanks{Equal contribution.} \quad Hanshu Yan\\ \quad Weiwei Lv \quad Yizhe Zhu \quad Jiashi Feng \\
  \\
  ByteDance Inc. \\
  \texttt{\{{daquanzhou, weimin.wang, hanshu.yan\}@bytedance.com}} \\
  \texttt{\{{vici, yizhe.zhu, jshfeng\}@bytedance.com}} \\
  }

\maketitle
\ificcvfinal\thispagestyle{empty}\fi

\begin{abstract}
   We present an efficient text-to-video generation framework based on latent diffusion models, termed \emph{MagicVideo}. MagicVideo can generate smooth video clips that are concordant with the given text descriptions. Due to a novel and efficient 3D U-Net design and modeling video distributions in a low-dimensional space, MagicVideo can synthesize video clips with  256$\times$256 spatial resolution on a single GPU card, which takes around 64$\times$ fewer computations than the Video Diffusion Models (VDM) in terms of FLOPs. 
   In specific, unlike existing works that directly train video models in the RGB space, we use a pre-trained VAE to map video clips into a low-dimensional latent space and learn the distribution of videos' latent codes via a diffusion model. 
   Besides, we introduce two new designs to adapt the U-Net denoiser trained on image tasks to video data: a frame-wise lightweight adaptor for the image-to-video distribution adjustment and a directed temporal attention module to capture temporal dependencies across frames. Thus, we can exploit the informative weights of convolution operators from a text-to-image model for accelerating video training.
   To ameliorate the pixel dithering in the generated videos, we also propose a novel VideoVAE auto-encoder for better RGB reconstruction. 
   We conduct extensive experiments and demonstrate that MagicVideo can generate high-quality video clips with either realistic or imaginary content. The code will be made public.
\end{abstract}


\section{Introduction}
\label{sec:intro}

Diffusion-based generative models have shown astonishing achievements over a variety of applications, including text-to-image generation \cite{imagen, dalle2,rombach_high-resolution_2022} and text-to-3D object generation \cite{poole2022dreamfusion}, 
due to their superior generation quality and scaling capability to large datasets.   For example, \mbox{DALL-E 2}~\cite{dalle2}, Imagen \cite{imagen} and Latent  Diffusion \cite{rombach_high-resolution_2022} models
can generate photo-realistic image contents from the given texts after being trained with large-scale image-text datasets (\eg, LAION 400M \cite{laion5b}).

\begin{figure*}
\centering
\setlength{\tabcolsep}{2pt}
    \makebox[\textwidth][c]{
    \begin{tabular}{ccccc}
    
         \includegraphics[width=0.17\textwidth]{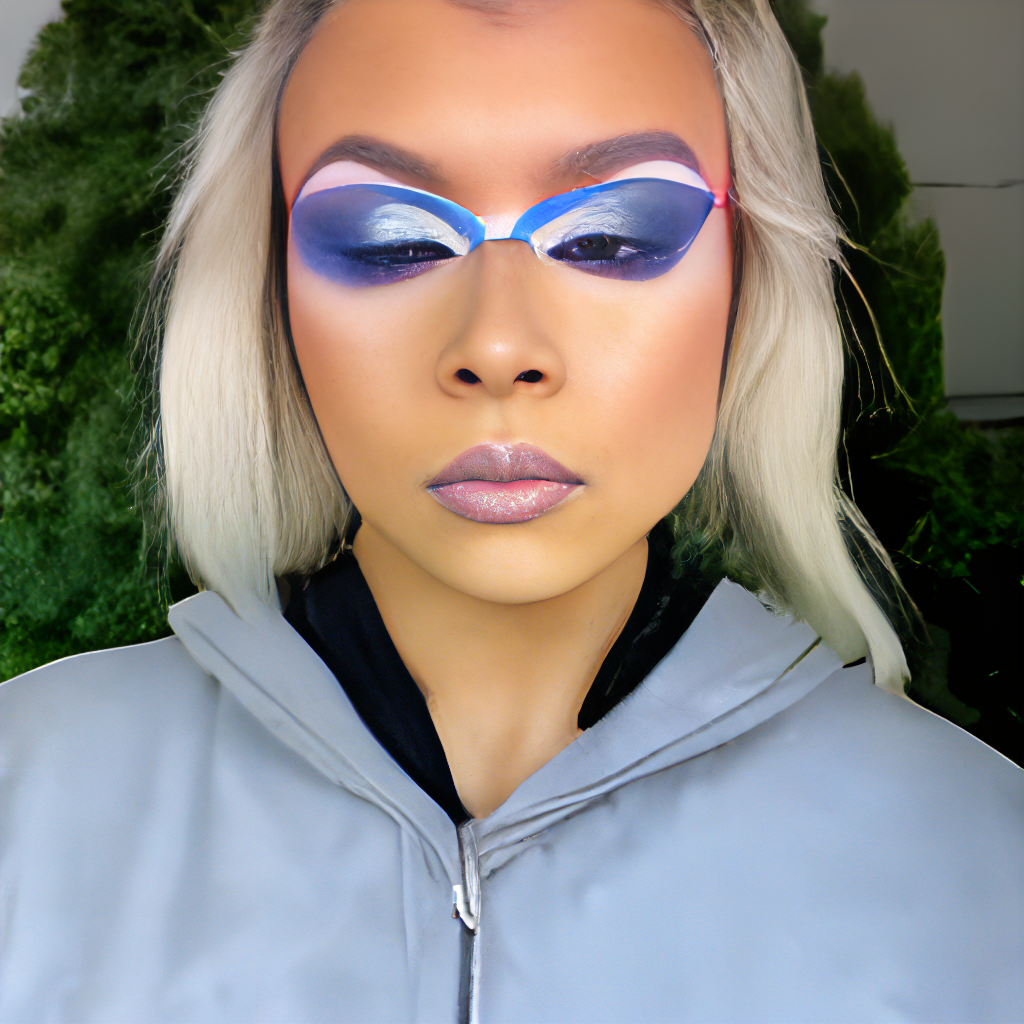} &
         \includegraphics[width=0.17\textwidth]{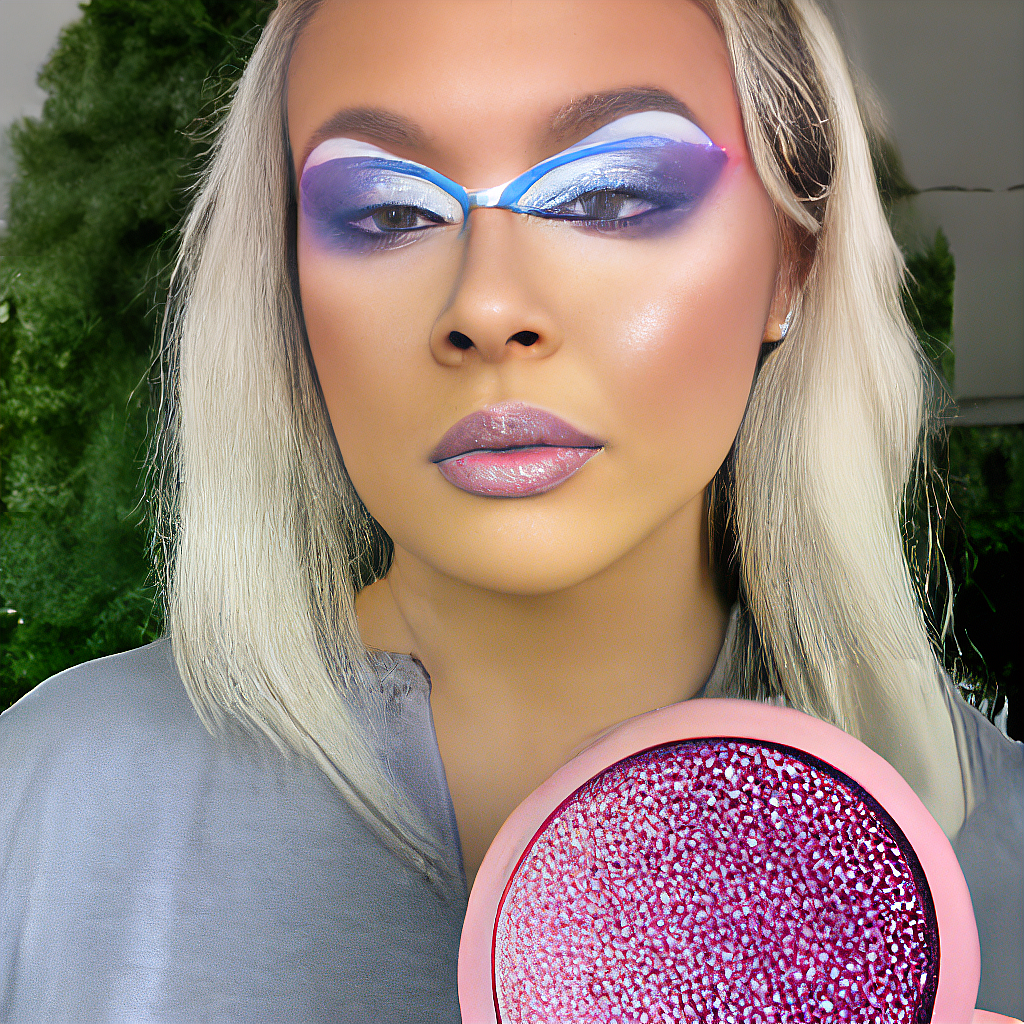} &
         \includegraphics[width=0.17\textwidth]{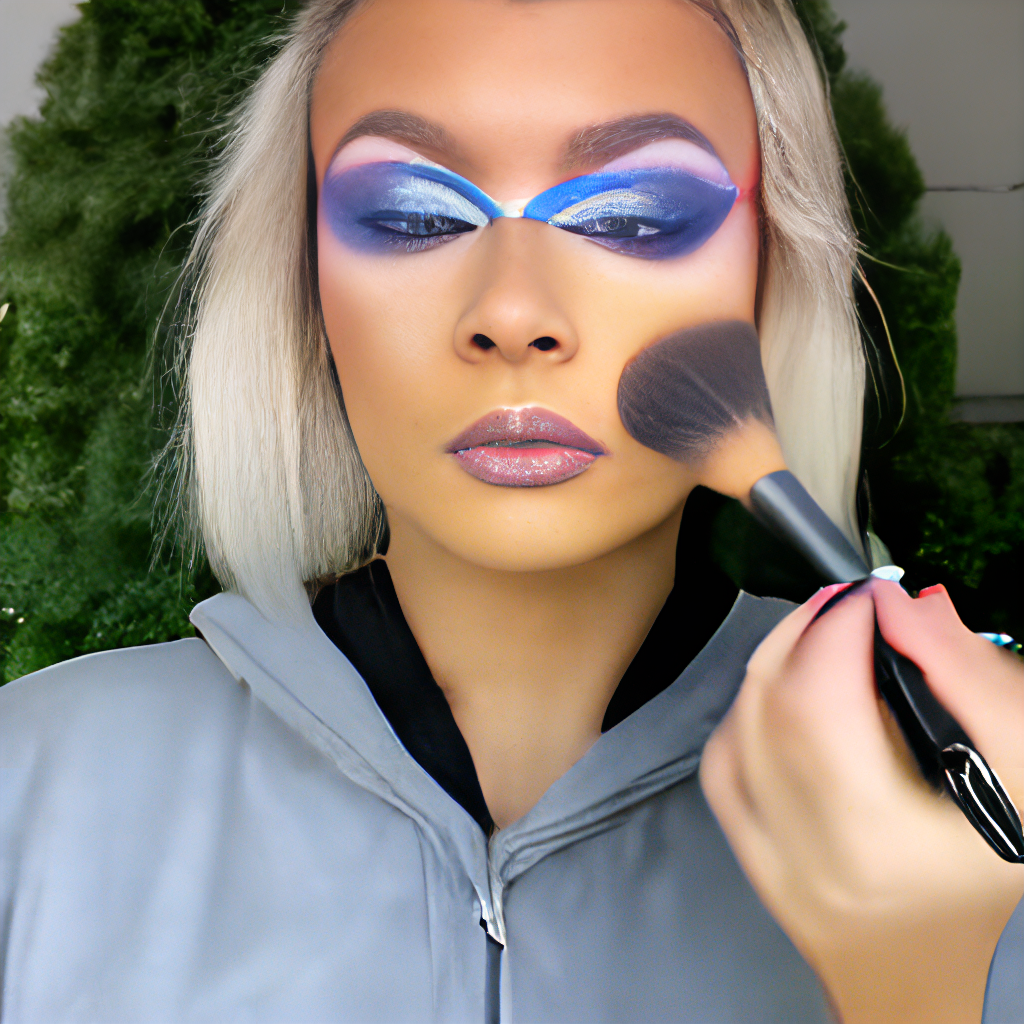} &
         \includegraphics[width=0.17\textwidth]{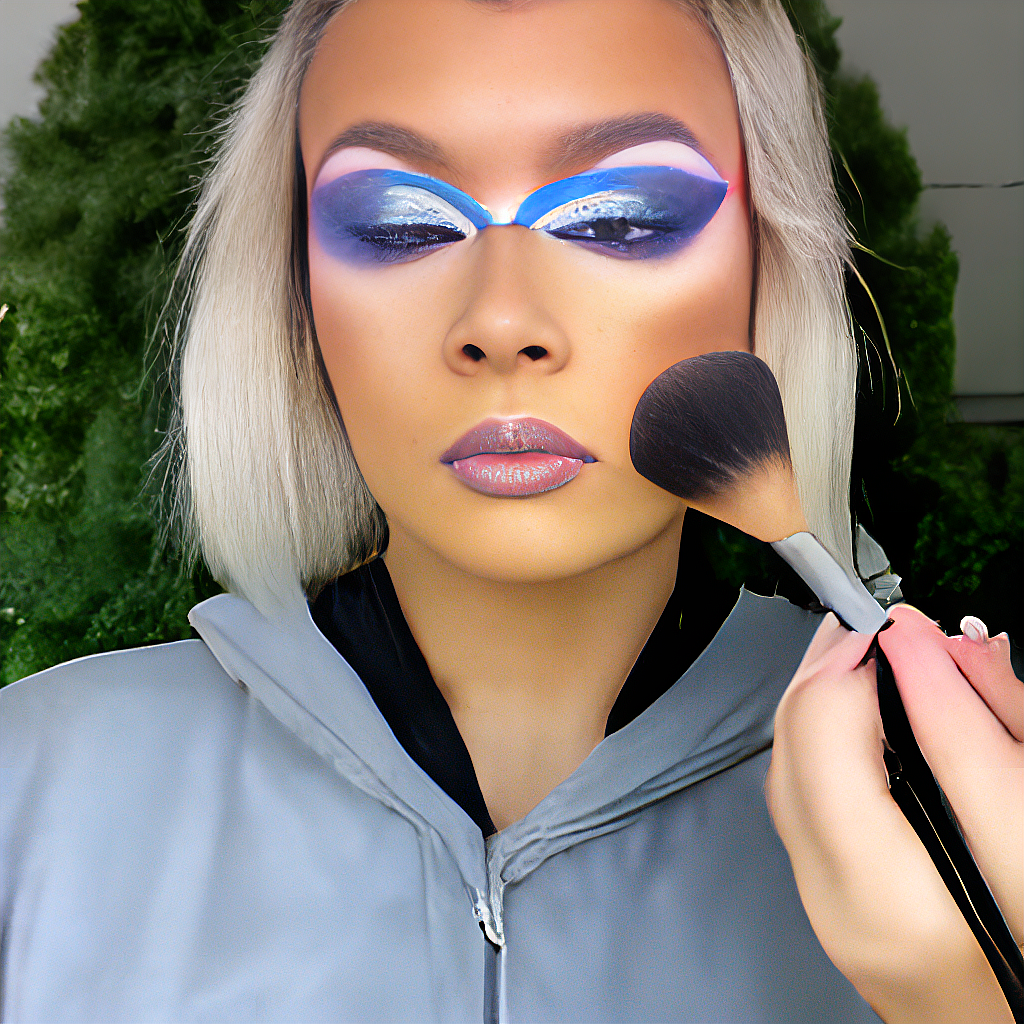} &
         \includegraphics[width=0.17\textwidth]{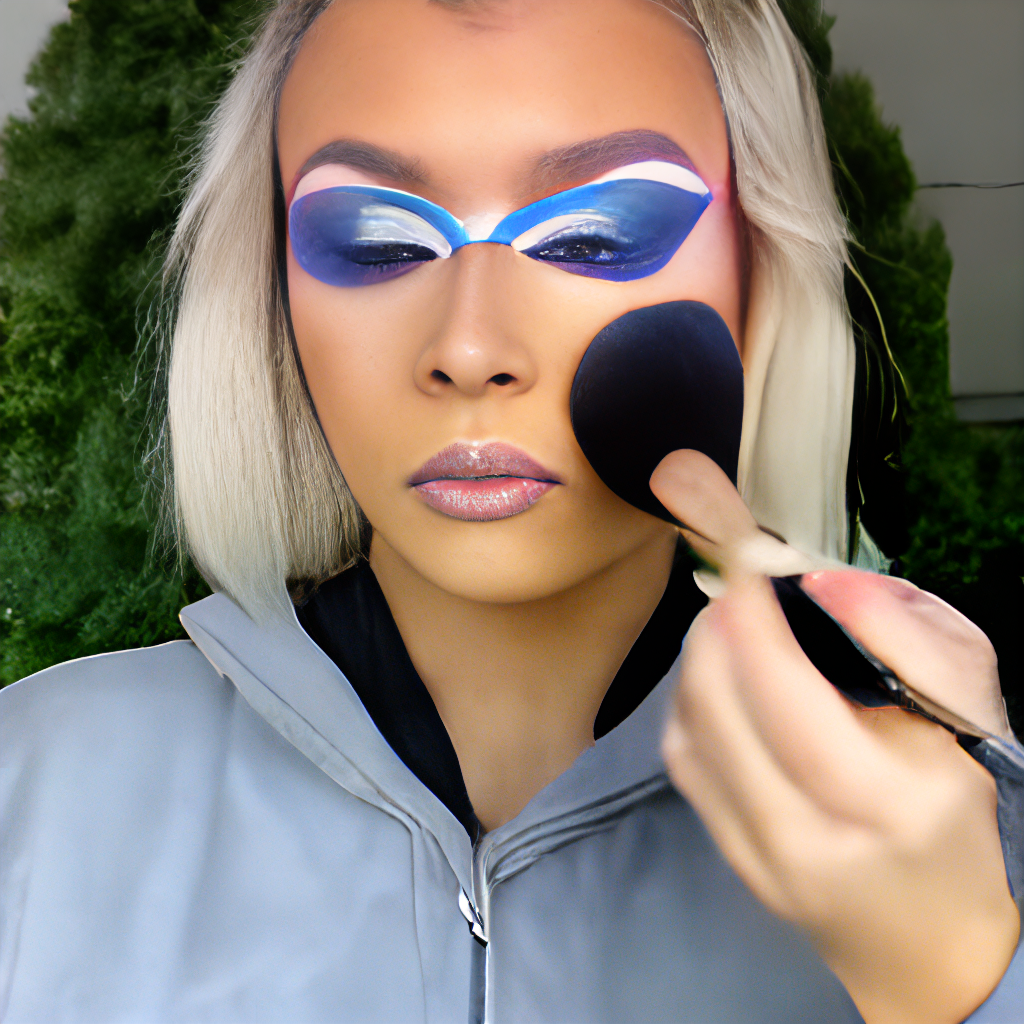} \\
         \multicolumn{5}{c}{\footnotesize Young attractive woman makeup in the morning.} \\[3pt]
    
         \includegraphics[width=0.17\textwidth]{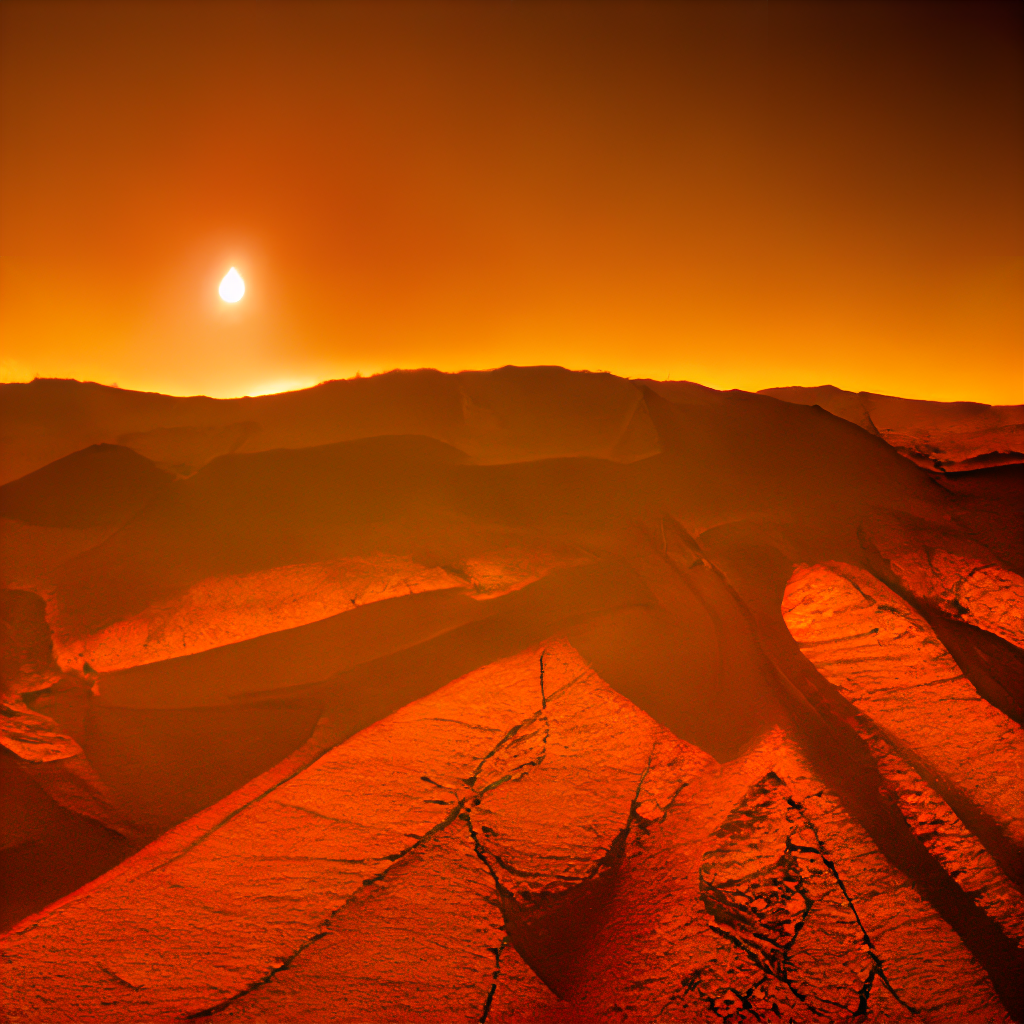} &
         \includegraphics[width=0.17\textwidth]{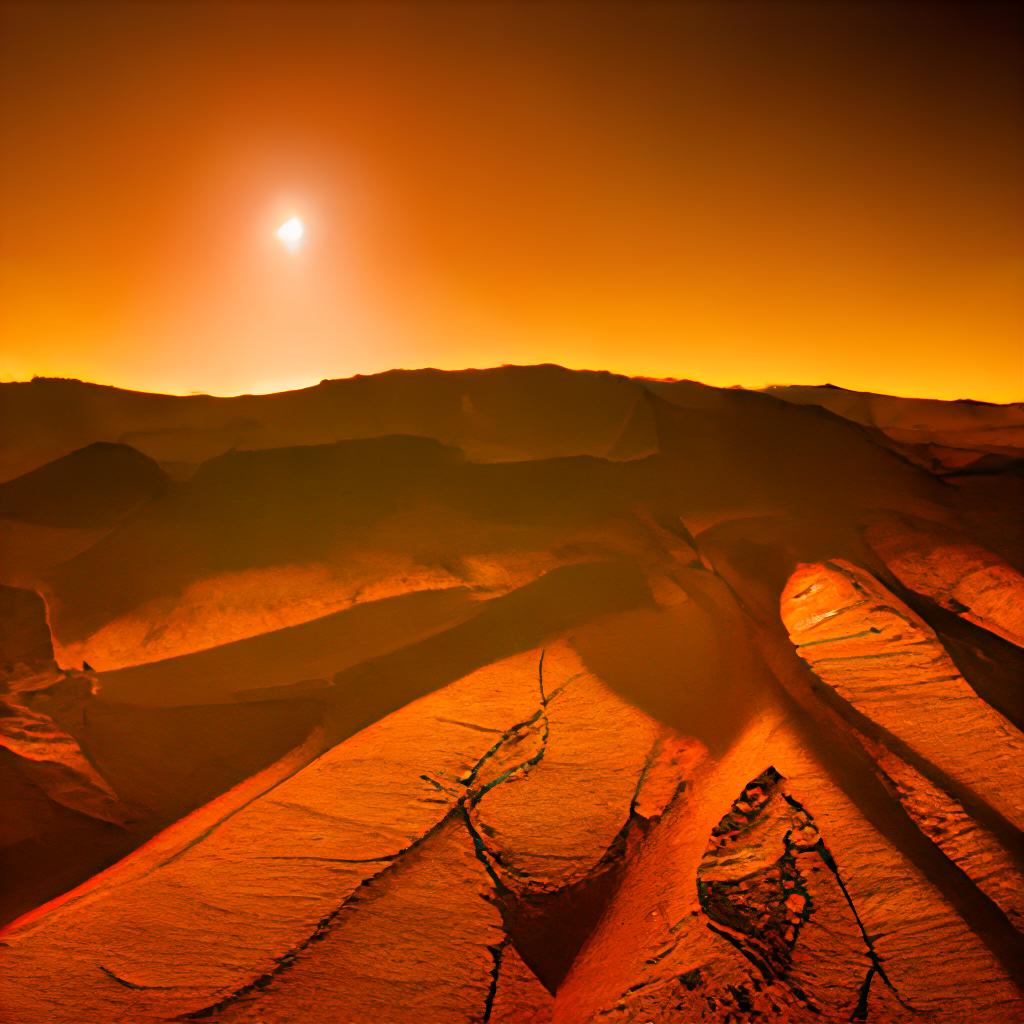} &
         \includegraphics[width=0.17\textwidth]{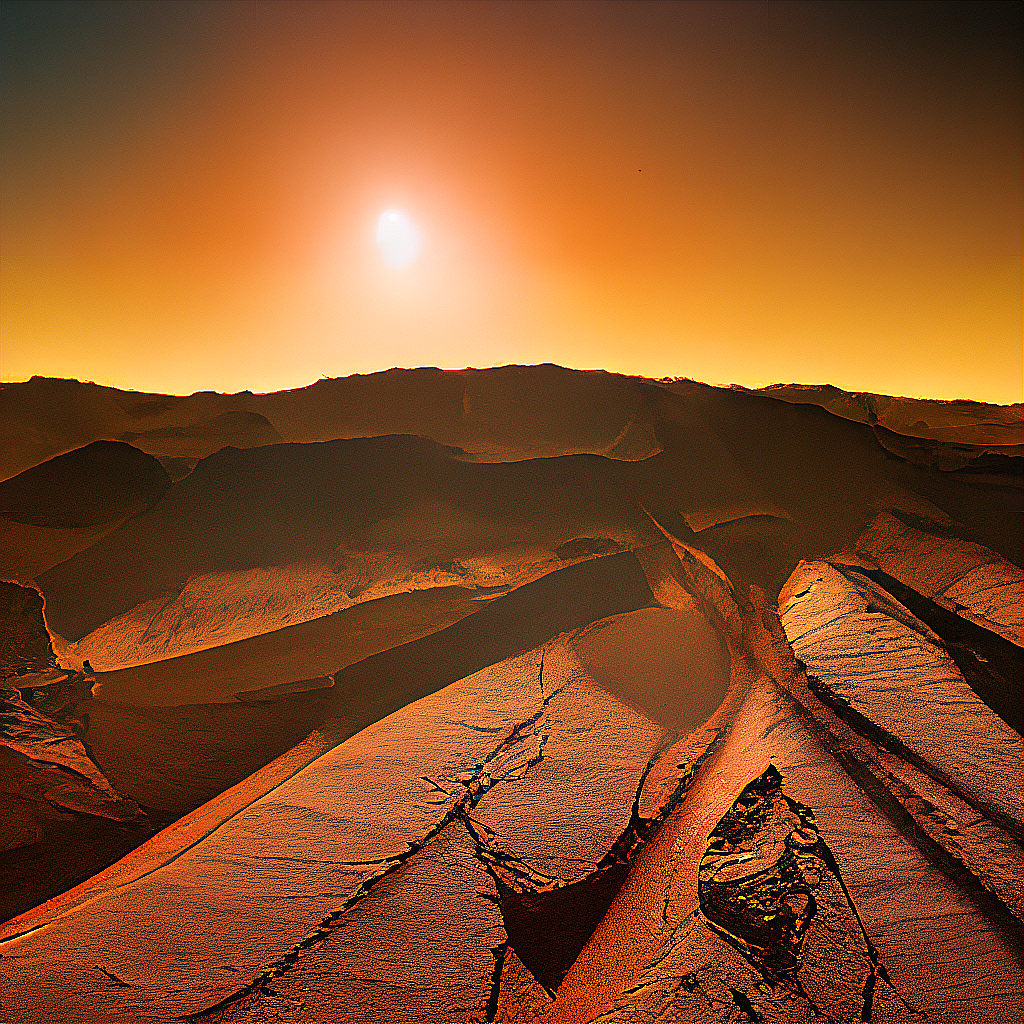} &
         \includegraphics[width=0.17\textwidth]{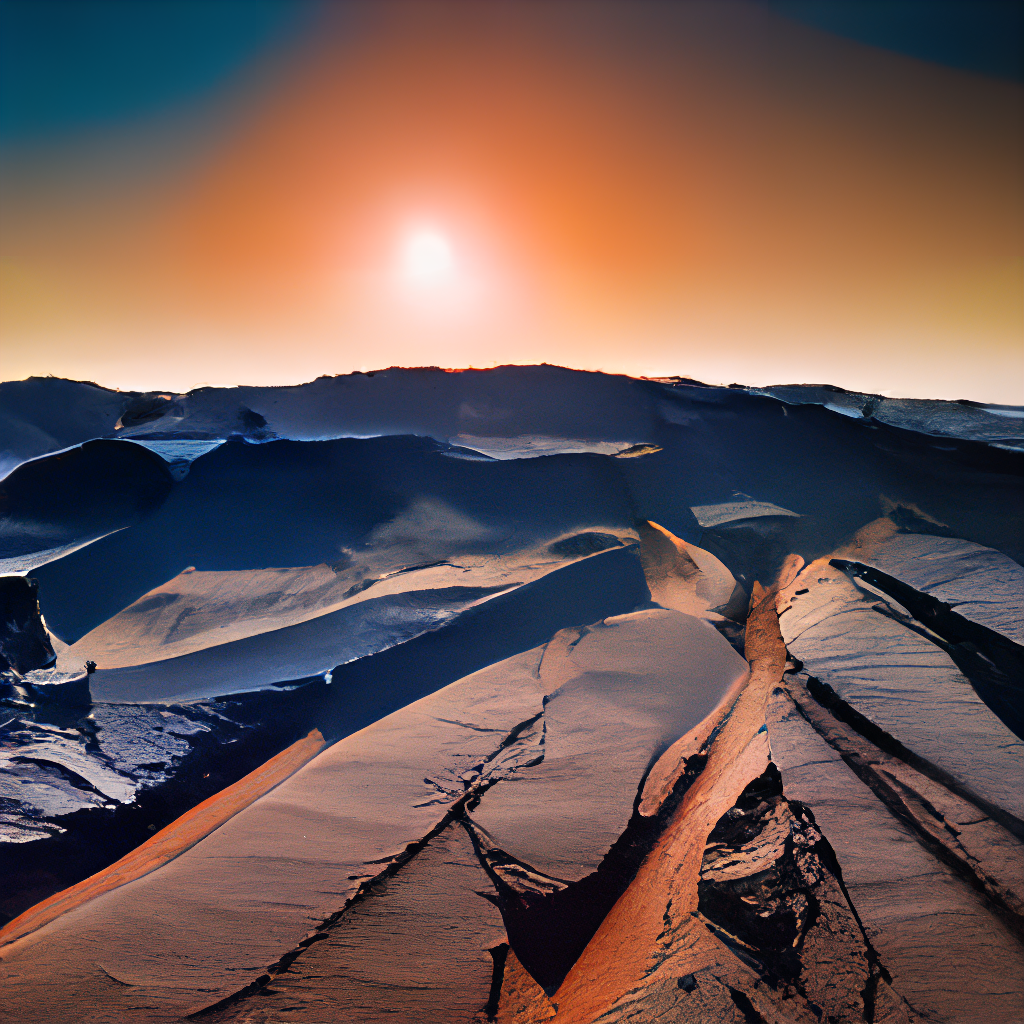} &
         \includegraphics[width=0.17\textwidth]{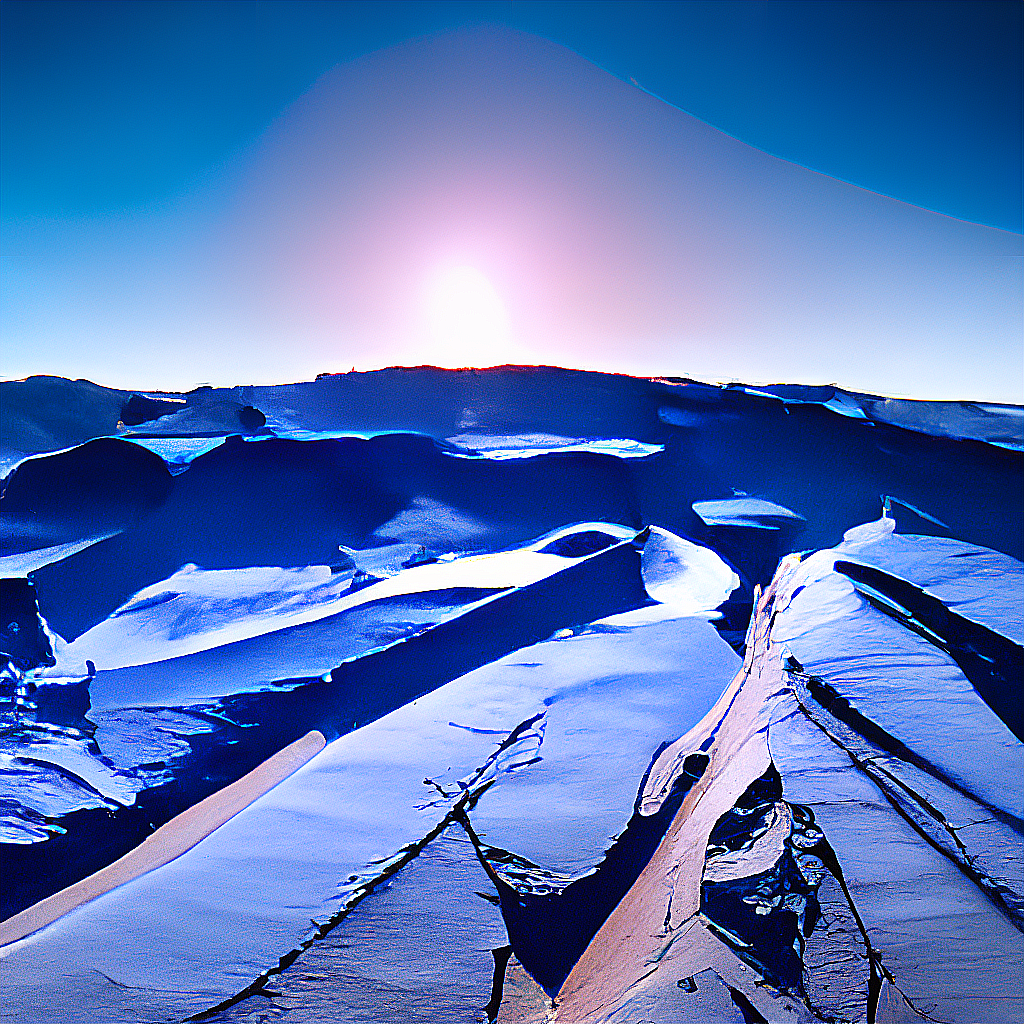} \\
         \multicolumn{5}{c}{\footnotesize A beautiful sunrise on mars, Curiosity rover. High definition, timelapse,dramatic colors.} \\[3pt]
         
         
         \includegraphics[width=0.17\textwidth]{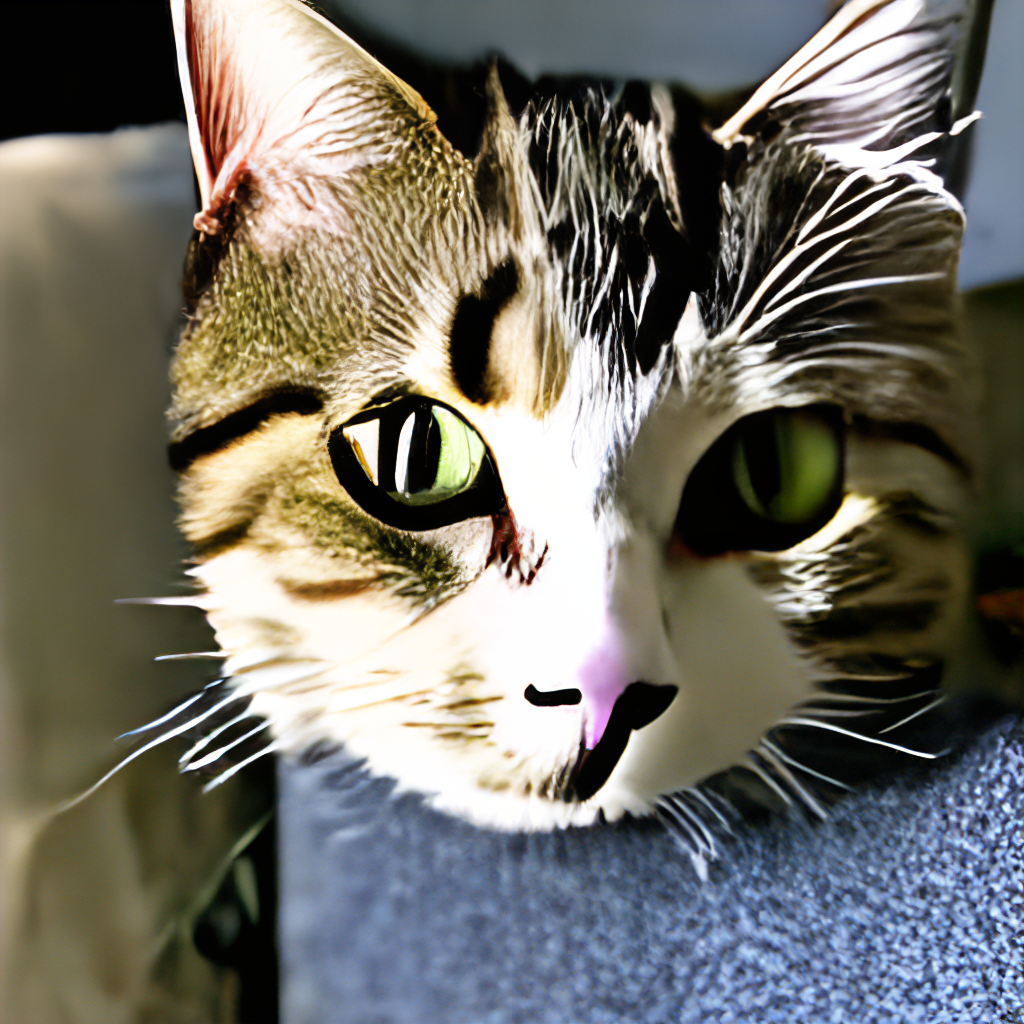} &
         \includegraphics[width=0.17\textwidth]{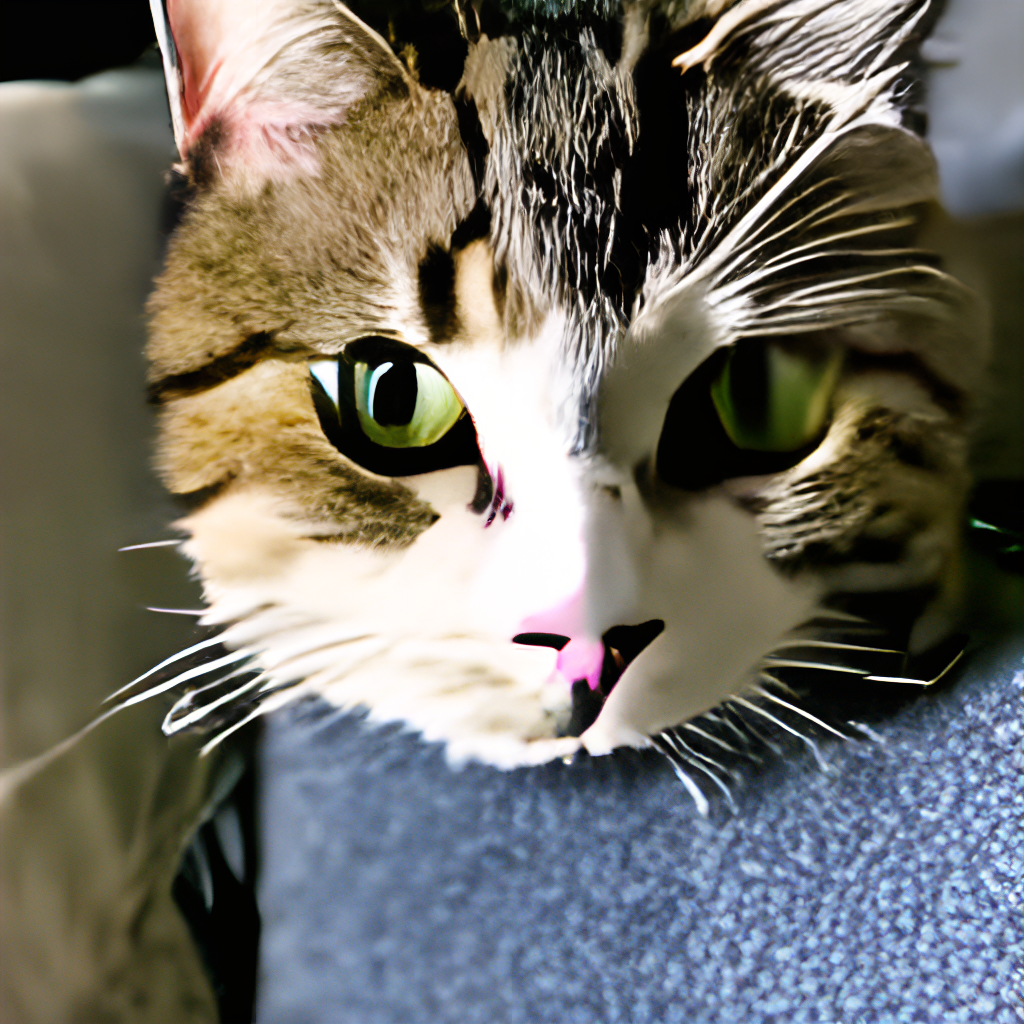} &
         \includegraphics[width=0.17\textwidth]{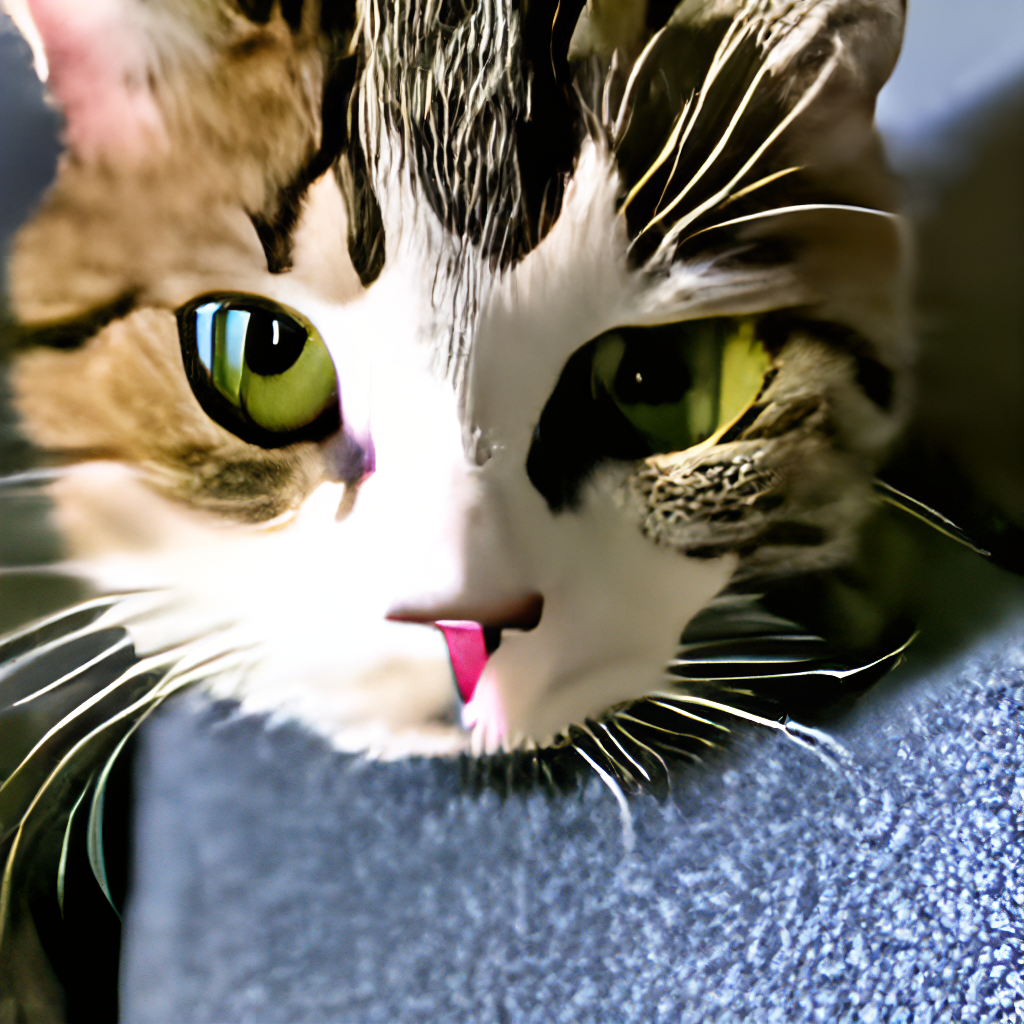} &
         \includegraphics[width=0.17\textwidth]{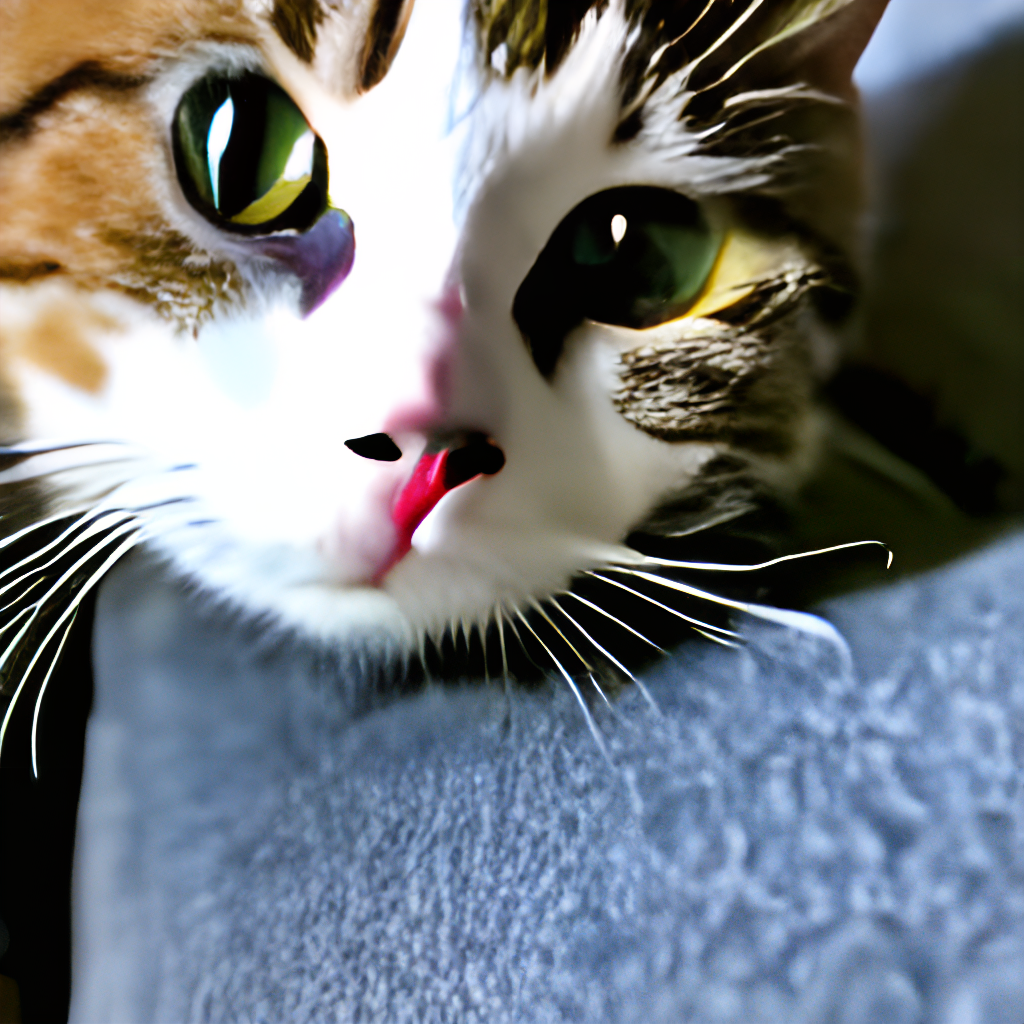} &
         \includegraphics[width=0.17\textwidth]{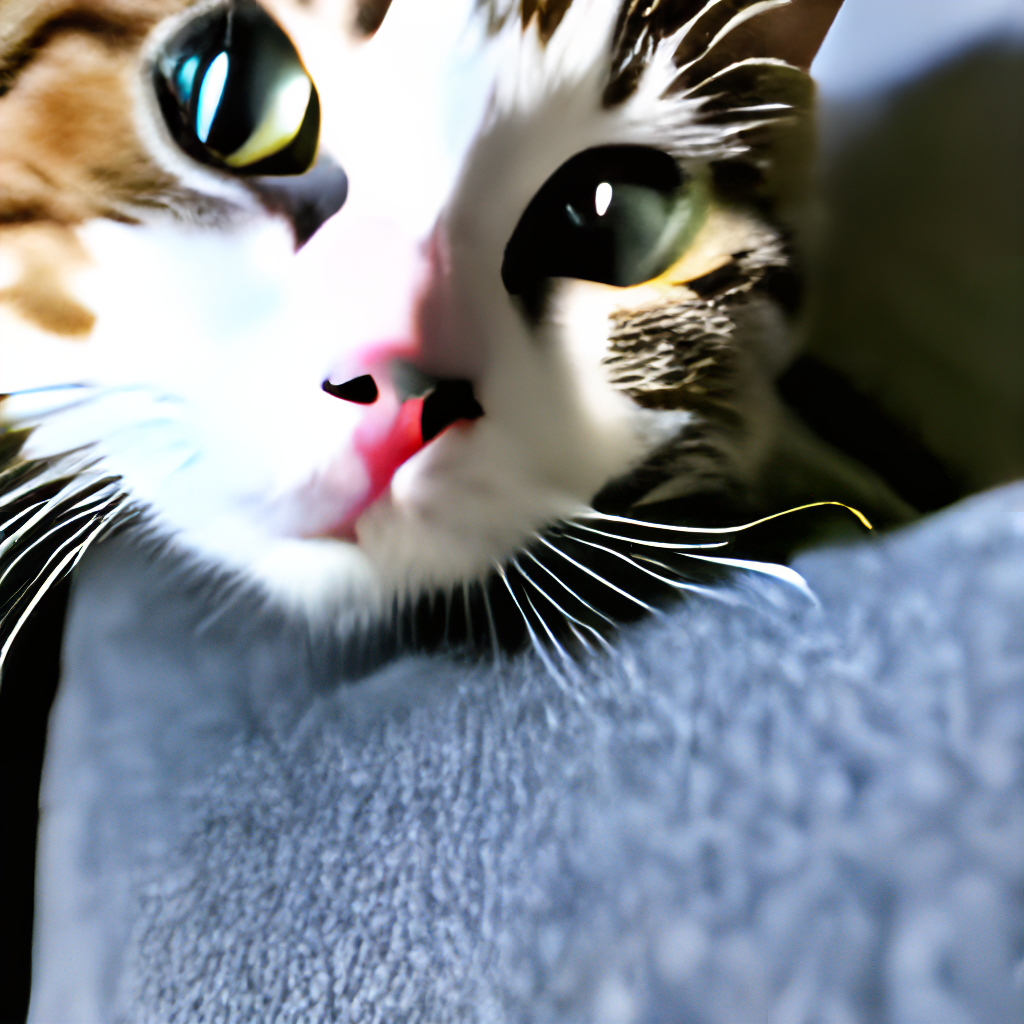} \\
         \multicolumn{5}{c}{\footnotesize A cute cat.} \\[3pt]
         
         \includegraphics[width=0.17\textwidth]{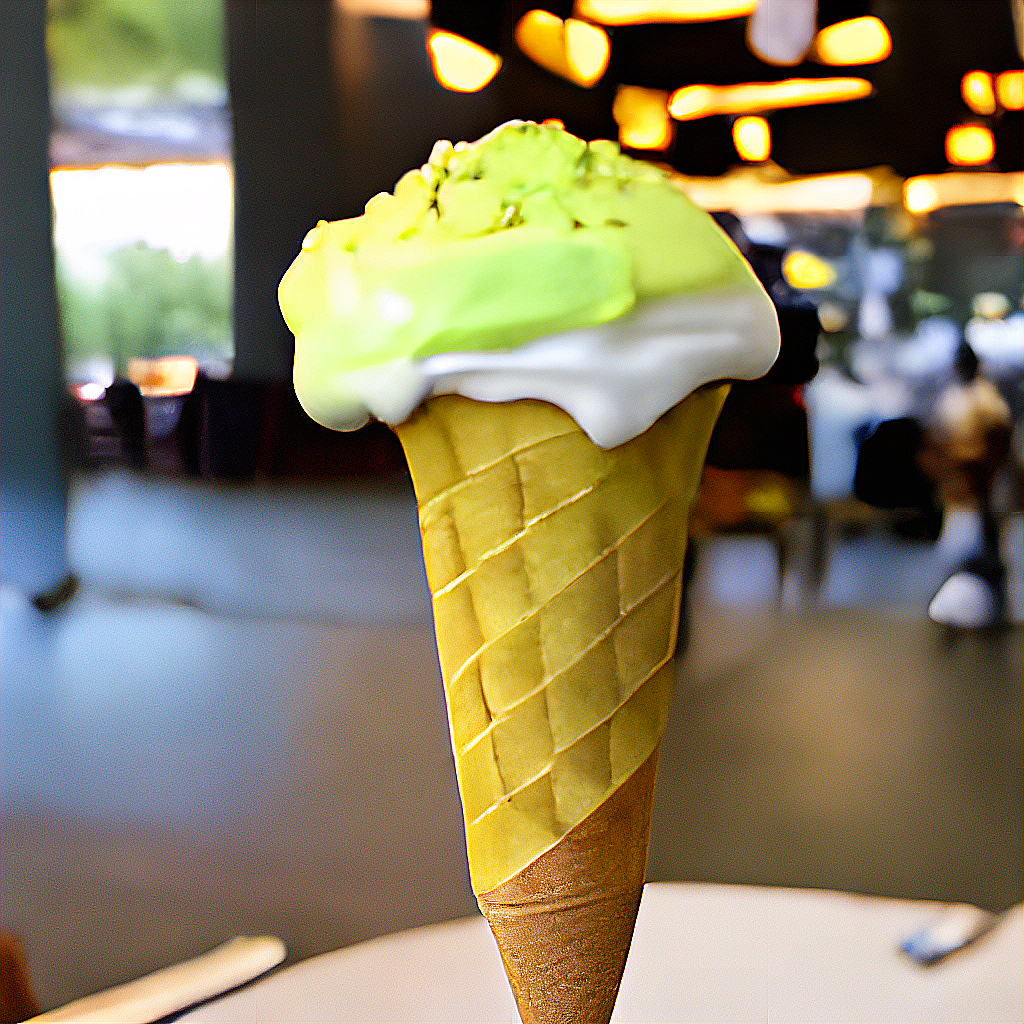} &
         \includegraphics[width=0.17\textwidth]{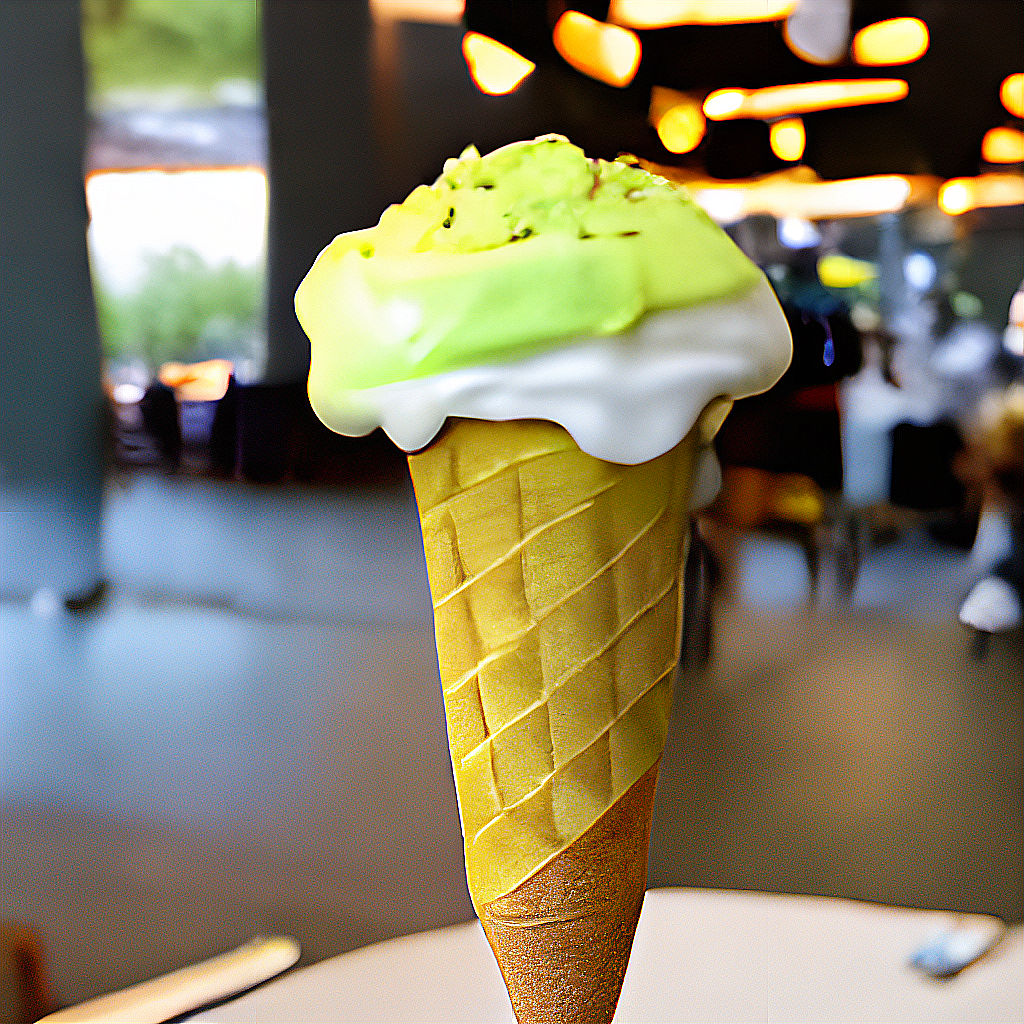} &
         \includegraphics[width=0.17\textwidth]{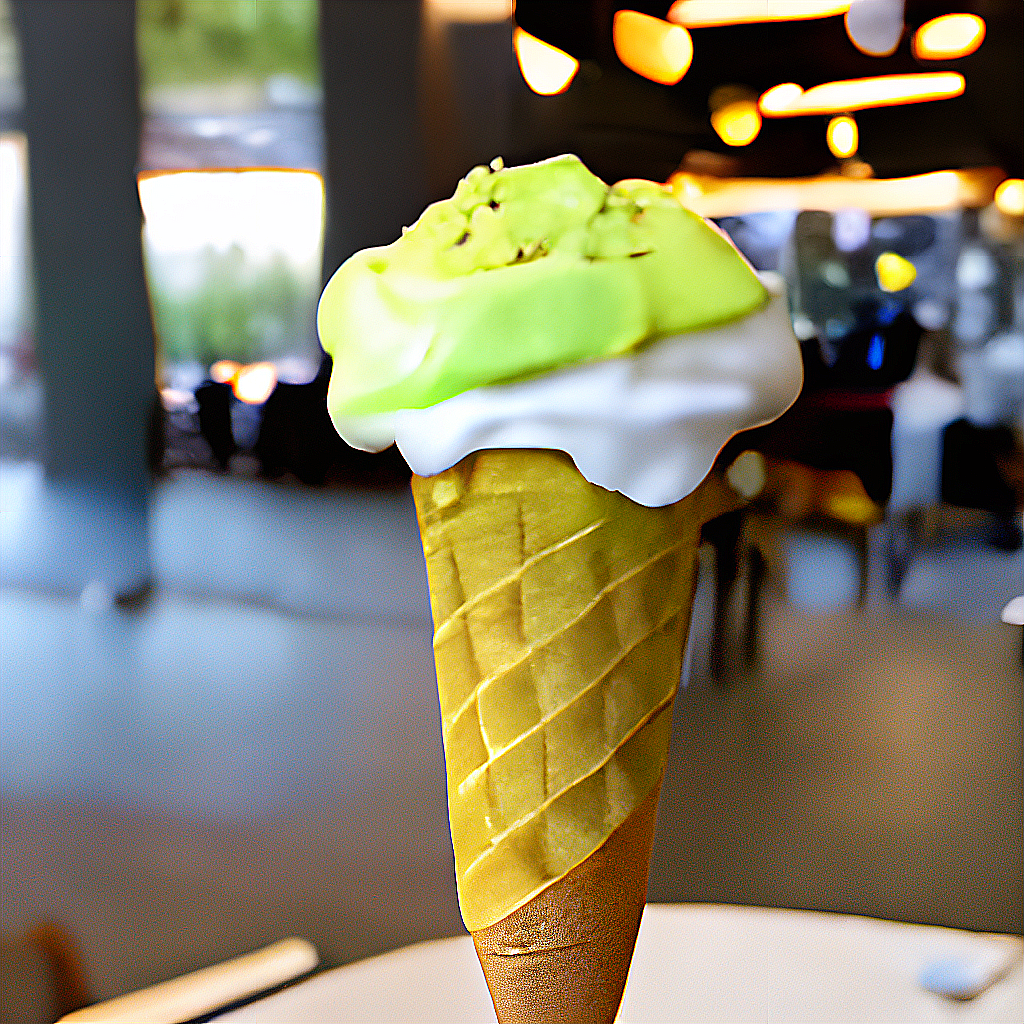} &
         \includegraphics[width=0.17\textwidth]{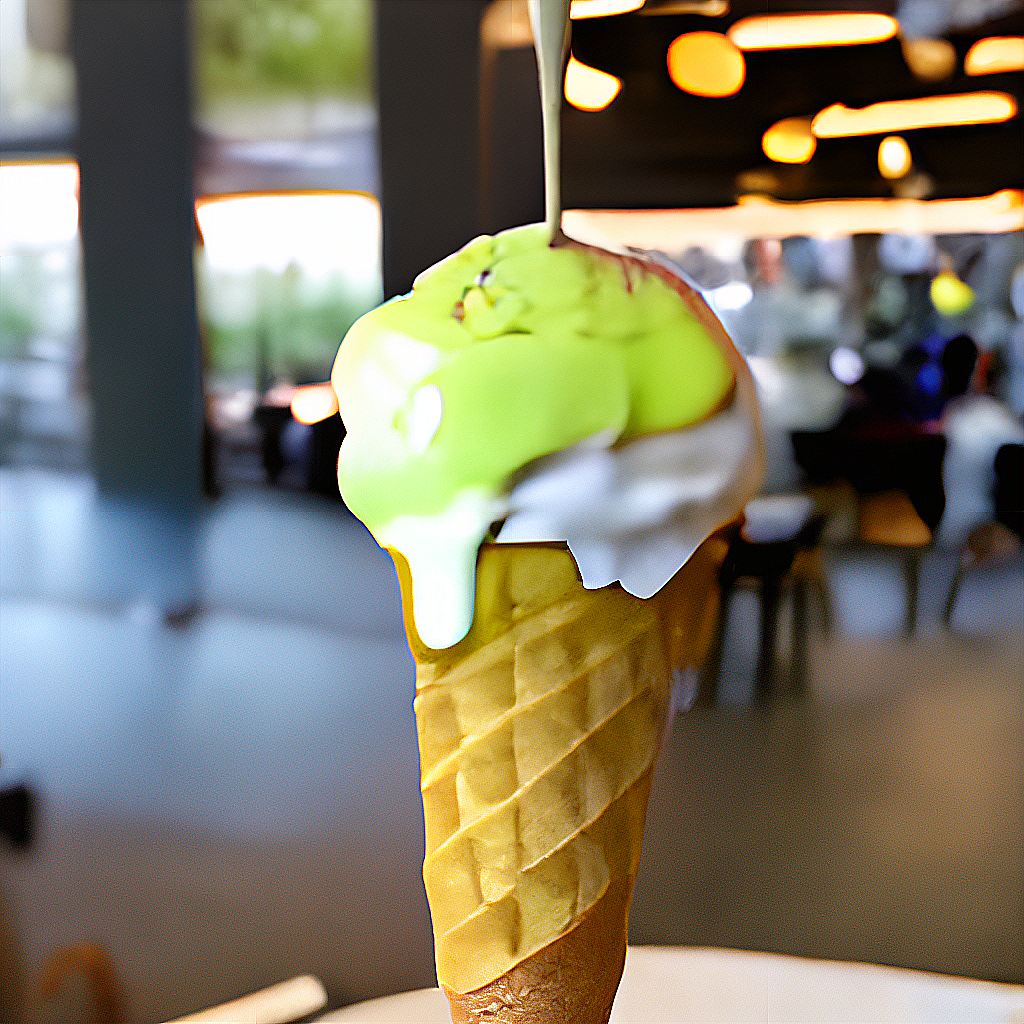} &
         \includegraphics[width=0.17\textwidth]{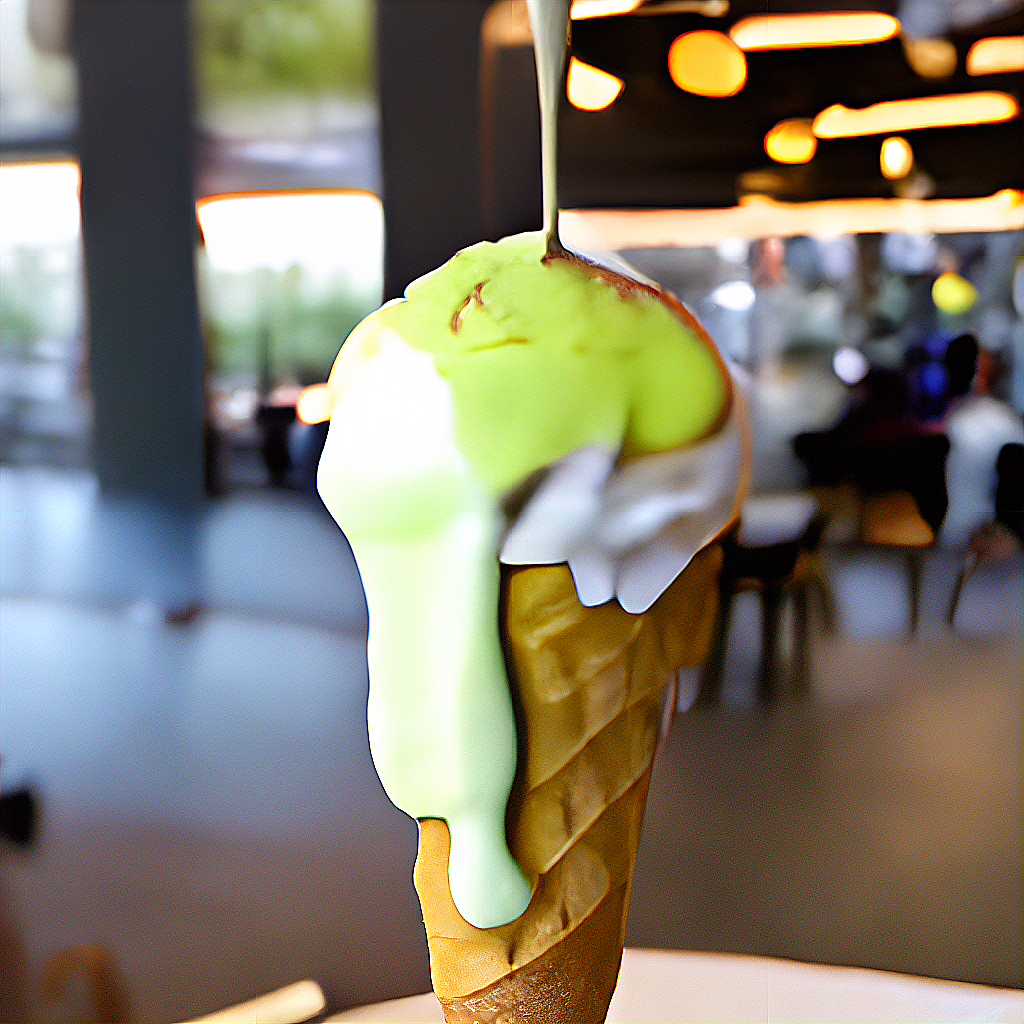} \\
         \multicolumn{5}{c}{\footnotesize Melting pistachio ice cream dripping down the cone.} \\[3pt]
         

    \end{tabular}}
    \caption{Given various text prompts, MagicVideo produces diverse and temporally-coherent videos that are well-aligned with the prompts. Note that the videos are generated with the key frame model directly without using super-resolution.}
    \vspace{-4mm}
    \label{fig:big_sample_figure}
\end{figure*}

Despite the success in text-to-image generation tasks, using diffusion-based generative models for video generation tasks  is still under-explored due to the following difficulties. \textit{1) Data scarcity}. Video data with precise textual descriptions are much harder to collect than image-text data, as videos are more difficult to   describe by a single sentence. Besides, different from images carrying compact information, each video may contain some redundant short clips that are   less relevant to the textual description. Such information redundancy  would limit     the effectiveness  of video data for model training. \textit{2) Complex temporal dynamics}. Video data contains diverse visual contents across frames and complex temporal dynamics. Therefore, it is much more challenging to model video data distribution compared to static images. \textit{3) High computation cost}. A smooth and informative video may contain more than hundreds of   frames. Compared to image generation, directly generating a whole video would consume  a huge amount of computational and memory resources.

Existing diffusion-based video generation models propose a cascaded pipeline~\cite{ho2022video} to deal with the high computational cost. This pipeline generates low-resolution video frames first via an iterative diffusion-based denoising process and then up-samples them  by a super-resolution module. Nevertheless, their computational cost  is still very high. For example, when generating a coarse  video clip of $16$ frames and $64\times 64$ resolution, the recent video diffusion model \cite{ho2022video}  would take 6-10 seconds with 75G GPU memory\footnote{We measure the speed on a single Nvidia A100 GPU card} for each diffusion iteration. The whole   generation process requires tens to hundreds of such  iterations for synthesizing a single frame,  causing unaffordable time cost.  

To further \textit{reduce the computational cost of video modeling}, we alternatively explore using the latent diffusion model (LDM) \cite{rombach_high-resolution_2022}  to learn the distribution of video data, which was developed   for image generation and   has shown state-of-the-art efficiency in generating images. Specifically, LDM first trains a variational auto-encoder (VAE)~\cite{rezende2014stochastic, razavi2019generating} to map images into a lower dimensional latent space. Then, it trains a diffusion model to approximate the distributions of the image's latent features instead of the raw RGB images. 
In this way, the spatial dimension of the latent features undergoing the diffusion-based denoising process is kept low,  thus reducing the computational cost significantly. For example, using VAE to reduce  the   frame resolution by 8 times, 
the computational cost would be reduced by around $64\times$ for every single frame generation. Motivated by this, we propose adopting the LDM to build our video generation model, MagicVideo.

To address other aforementioned challenges   for video generation, including \tit{data scarcity and complex temporal dynamics modeling},  we introduce  the following novel designs  to build the first LDM-based video generation model. To \tit{improve data efficiency} and alleviate the demand for paired video-text data,  we adopt 2D convolution  with temporal computation operators to model the spatial and temporal video features  instead of building the model  with vanilla 3D \cite{ho_video_2022} or (2+1)D \cite{singer_make--video_2022} convolutions.  This new architecture design allows for initializing 2D convolutions with the parameters of a pre-trained text-to-image model (\eg, LDM~\cite{rombach_high-resolution_2022}) and exploiting its prior knowledge of image generation for facilitating video modeling. Experiments demonstrate that this strategy enables the learning of video generation even with little video training data. To further reduce the memory cost, we use the same 2D convolutions for generating each frame. However, to avoid  deteriorating the temporal consistency in the generated keyframes (\eg, object motion),    we introduce a new and lightweight \textit{adaptor} module to adjust the feature distribution per frame. The adaptor only consists of a few scalar parameters yet performs well as it exploits the   correlation of  the video frames. It effectively alleviates the need for  independent    2D convolution blocks for modeling different frames \cite{lin2022frozen}. 

To   model the  \textit{temporal dynamics}, our model leverages the \textit{directed self-attention} mechanism. It calculates  the features of the future frame    based on all the preceding frames while keeping the previous frames   unaffected by the future ones. This improves the motion consistency over conventional bi-directional self-attention modules that existing generative models widely use \cite{singer_make--video_2022, imagen, rombach_high-resolution_2022, dalle2}. Furthermore,  we propose a novel VideoVAE containing a decoder block that is dedicated to reducing frame generation artifacts (\eg, pixel dithering). Different from the original VAE used in SD that treats each frame independently, VideoVAE considers the temporal relations during the decoding phase, leading to a more consistent high-frequency content.
%

We conduct extensive experiments to verify the effectiveness of MagicVideo in generating high-resolution videos.   MagicVideo can generate photo-realistic video frames with  smooth motion and   consistent object identity, as shown in Fig.~\ref{fig:big_sample_figure}, achieving higher quality and efficiency than recent strong methods.  We also present its applications to image-to-video and video-to-video generations  to show  its versatility for various conditional video generation schemes.  


\section{Related works}
\myPara{Diffusion based generative models.} Denoising diffusion probabilistic models (DDPMs)  have achieved great success in image generation~\cite{ho_denoising_2020, song_score-based_2021} and editing \cite{lugmayr_repaint_nodate, meng_sdedit_2022, hertz_prompt--prompt_2022, liew_magicmix_2022, kawar_imagic_2022}. For example, DALL-E-2~\cite{dalle2} uses a generative model (\eg autoregressive or DDPM) to learn the distribution of images' CLIP embeddings and then train a DDPM to synthesize RGB images by conditioning on the sampled embeddings. Alternatively, Imagen~\cite{imagen} directly models the distribution of low-dimensional RGB images via a DDPM and uses cascaded super-resolution models to   enhance image qualities. 
However, due to the intrinsic   iterative sampling, the computational overhead of DDPMs gets very high and impedes their applications. To improve the efficiency, Rombach \etal~2022~\cite{rombach_high-resolution_2022} proposed the latent diffusion model (LDM) that models the data distribution in a low-dimensional latent space. 
Denoising noisy data in a lower dimension may reduce the computational cost in the generation process. 
Specifically, LDM first trains an autoencoder  to map images  into a low-dimensional space and reconstruct images from latent features. Then, a DDPM with a time-conditional U-Net backbone is used to model the distribution of the latent representations.

\myPara{Video generation.} Various video generation methods have been proposed in the past, including   GAN-based methods~\cite{vondrick_generating_2016, clark_adversarial_2019, tulyakov_mocogan_2017} and auto-regressive one~\cite{ranzato_video_2016, kalchbrenner_video_2017, babaeizadeh_fitvid_2021, kumar_videoflow_2020, mathieu_deep_2016, hong_cogvideo_2022}. Recently, the success of diffusion-based   models for image generation also triggered significant interest in exploring their applications in video modeling. For \textit{unconditional video generation}, Ho \etal  \cite{ho_video_2022} extended image DDPM models to the video domain by developing a 3D U-Net architecture. Harvey \etal~\cite{harvey_flexible_2022} proposed to model the distribution of subsequent video frames   in an auto-regressive manner. In this work, we are interested in synthesizing videos in a controllable manner, i.e., \textit{text-conditional video generation}.
Along this line, Hong \etal~\cite{CogVideo} proposed an auto-regressive framework, CogVideo, that models the video sequence by conditioning  itself for the given text and the previous frames.  Ho \etal \cite{ho_imagen_2022} proposed a diffusion-based cascaded pipeline, Imagen Video that consists of one base text-to-video module and three spatial and temporal super-resolution modules. 
Concurrently, Singer \etal  \cite{singer_make--video_2022} propose a multi-stage text-to-video generation method, termed Make-A-Video, that first exploits a text-to-image model to generate image embeddings and then trains a low-resolution video generation model with conditioning on the image embeddings, which are then up-sampled via   super-resolution models. Both Imagen Video and Make-A-Video model the video distribution in the RGB space. Differently, we explore a   more efficient way for video generation by synthesizing videos in a low-dimensional latent space.

\section{Method}
\begin{figure*}[t] 
\centering
\includegraphics[width=1.0\linewidth]{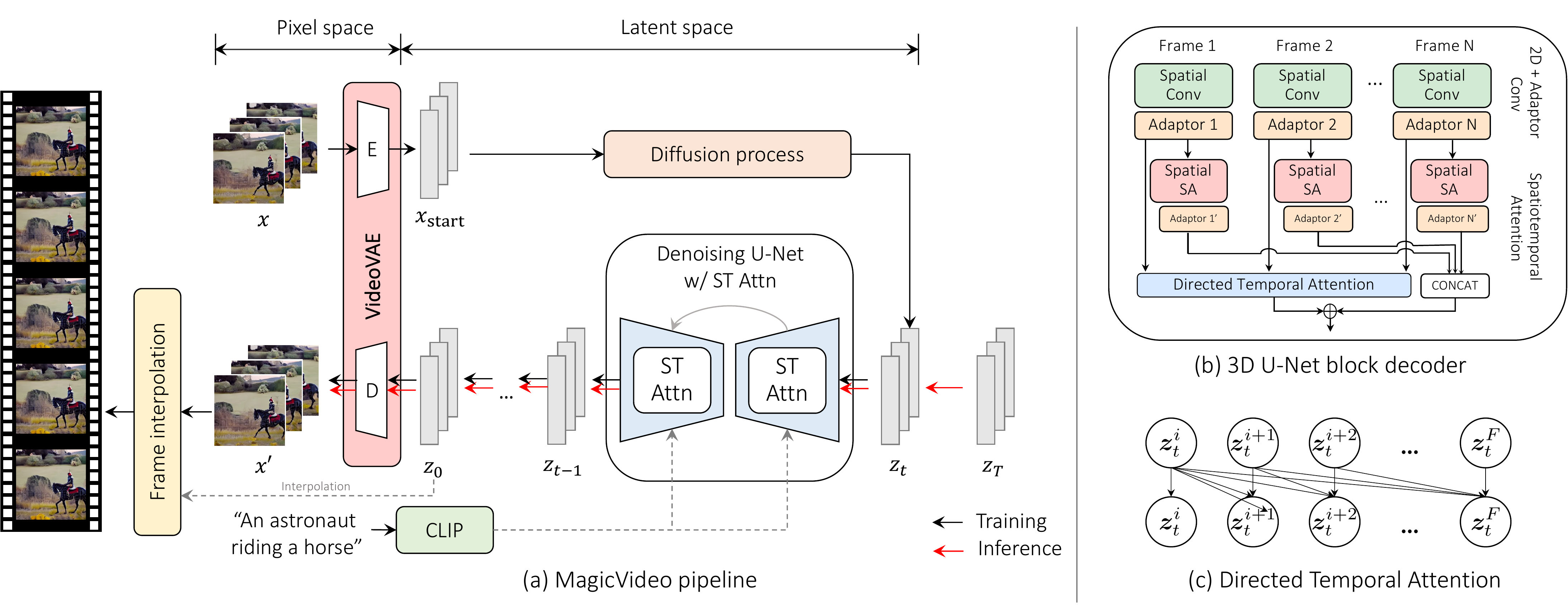}
\caption{\textbf{The overall framework of MagicVideo.} (a) The data flow of both the training and inference phases: during the training phase, a timestep $t$ will be sampled randomly from $[0, T]$ and the input video frames are corrupted via the diffusion process, and a U-Net decoder is used to learn to reconstruct the video frames. Gaussian noise is randomly sampled during inference, and the denoising process is repeated $T$ times. The denoised latent vector $\vz$ is then fed into a VAE decoder and converted to the RGB space. (b) The structure of the spatiotemporal attention (ST-Attn) module. (c) The directed attention used in the ST-Attn. For the details of the VideoVAE decoder design, please refer to Fig. \ref{fig:videovae}(b).}
\vspace{-3mm}
\label{fig:pipeline}
\end{figure*}

In this section, we introduce the MagicVideo in details (Fig.~\ref{fig:pipeline}). 
MagicVideo models video clip distribution in a low-dimension latent space. During the inference stage, MagicVideo first generates key frames in the latent space; then it interpolates key frames to smoothing the frame sequence temporally and maps the latent sequence back to RGB space. Finally, MagicVideo upsamples the obtained video to a high-resolution space for better visual quality.

\myPara{Notation.} In this paper, we use $\vx_t$ to denote a sequence of video frames corrupted with Gaussian noise at intermediate time step $t$. $\vx_t$ is short for $\vx_t = [\vx_t^1, ..., \vx_t^F]$, where $\vx_t^i$ represents the $i^{\text{th}}$ frame in the sequence. The encoder and decoder of the proposed video variational auto-encoder (VideoVAE) are denoted by $\mathcal{E}(\cdot)$ and $\mathcal{D}(\cdot)$, respectively. The video frames are mapped into the latent space one by one, \ie, $\vz_t = [\mathcal{E}(\vx_t^1), ..., \mathcal{E}(\vx_t^F)]$.
We use CLIP \cite{radford2021learning} to encode the given text prompt $\vy$, and the obtained embedding is denoted as $\tau(\vy)$. We use $\epsilon_{\theta}(\vz_t, t, \tau(\vy))$ to represent the denoiser of the diffusion model in the latent space.

\subsection{Key frame generation}
The most crucial step of MagicVideo is key frame generation. We use a diffusion model to approximate the distribution of {16} key frames in a low-dimensional latent space. 
In specific, we design a novel 3D U-Net decoder with an efficient video distribution adaptor and a directed temporal attention module, for video generation. We follow   LDM~\cite{rombach_high-resolution_2022} for  image generation to add text conditioning via cross-attention, where the text embeddings are used for computing the value and key embeddings, and the intermediate representations of U-Net are used for the query embeddings.

\subsubsection{Video distribution adaptor}
\label{subsubsec:adaptor}
The conventional operator in a neural network model for video data processing is the 3D convolution \cite{carreira2017quo}. However, the computation complexity   of 3D convolution is significantly higher than that of 2D convolution. Thus, to reduce the high computational cost, recent video processing models typically replace 3D convolution with a 2D convolution along the spatial dimension followed by a 1D convolution~\cite{tran2018closer} along the temporal dimension (termed ``2D+1D'').

In this work, we further simplify the operators  from ``2D+1D'' to ``2D+adaptor'', where the \textit{adaptor} is an even simpler operator compared to the 1D convolution. Specifically, given a sequence of $F$ video frames, we apply a {shared} 2D convolution for all the frames to extract their spatial features. After that, we adjust the mean and variance for the intermediate features of every single frame via: 
\begin{equation}
    \vz^i_t = S^i \cdot \text{Conv2d}(\vz^i_t) + B^i,
\end{equation}
where $\vz^i_t$ denotes the feature of the $i^{\text{th}}$ frame at denoising time step $t$, and $S, B \in \mathbb{R}^{F \times C}$   are two groups of learnable parameters. This operation is inspired by the observation that frames within each video clip are semantically similar. The small difference among frames may not be necessary for a dedicated 1D convolution layer. Instead, we model those differences via a small group of parameters. The details    of the adaptor  are shown in Fig.~\ref{fig:pipeline}(b).

\subsubsection{Spatial and  directed temporal attention}
\label{subsubsec:temp_attn}
Following previous works \cite{dalle2, imagen, ho_video_2022}, within the U-Net, we adopt self-attention modules after the down-sampling blocks that reduce the feature spatial resolution  by 4$\times$, 8$\times$  and 16$\times$. The attention operations are conducted along the spatial and temporal dimensions separately. The output of the two parallel attention modules is added and passed to the following modules:
\begin{equation}
\begin{multlined}
    \bm{z}_t = \text{S-Attn}(\bm{z}_{t-1}) + \text{T-Attn}(\bm{z}_{t-1}),
\end{multlined}
\end{equation}
 where S-Attn denotes the attention calculated along the spatial dimension (i.e., to aggregate the frame-wise feature tokens), and T-Attn denotes the self-attention conducted along the temporal dimension. Concretely, the spatial attention is calculated following previous works \cite{imagen, dalle2} via:
\begin{equation}
\text{S-Attn} = \text{Cross-Attn}(\text{LN}(\text{MHSA}(\text{LN}(\bm{z}_{t-1}))), \tau(\vy) ),
\end{equation}
where $\text{MHSA}$ is a standard Multi-head  Self-attention module used in vision transformers \cite{dosovitskiy2020image, zhou2022understanding}, LN denotes the layer normalization \cite{ba2016layer}, and cross-Attn denotes the cross self-attention module where the attention matrix is calculated between the frame tokens $\vz_{t-1}$ and the text embedding $\tau(\vy)$. Different from recent VDM \cite{ho_video_2022}, 
we introduce a novel directed self-attention module to better model the video temporal dynamics for the denoising decoder. 

\myPara{Directed temporal attention.}
Recent video generation frameworks \cite{singer_make--video_2022, ho_video_2022} mostly use a conventional (i.e., bi-directional) self-attention along the temporal dimension for the motion learning in the video dataset. We notice that the self-attention matrix missed a critical feature of the video data: the motions are directional. In videos, the frames are expected to change in a regular pattern along the temporal dimension. We propose a directed self-attention mechanism to inject the temporal dependency among the frames.

Given a set of given video frames features, $\vz_t \in \mathbb{R}^{F\times C\times H \times W}$ where $C, F, H, W$ denotes the batch size, number of feature channels, number of frames and the spatial dimension of the features respectively. We first reshape $\vz_t$ into shape $HW \times \textit{\#Heads} \times F \times \frac{C}{\textit{\#Heads}}$ and treat each pixel of each frame as a token, where $\textit{\#Heads}$  denotes the number of attention heads.
The temporal attention is applied to the tokens of the exact spatial location across different frames to model their dynamics. 
Specifically,  we obtain their query $Q_t$, key $K_t$, and value $V_t$ embeddings for self-attention via three linear transformations, then calculate the temporal attention  matrix, $A_t \in \mathbb{R}^{\textit{\#Heads}\times F \times F}$ via:
\begin{equation}
     A_t =  \text{Softmax}(Q_t K^\top_t/\sqrt{d}) \odot M, 
\end{equation}
where $d$ is the dimension of   embeddings per head and $M$ is an lower triangular  matrix with   $M_{p,q} = 0$  if $p > q$ else 1. With the   mask, the present token is only affected by the previous tokens and independent from the future tokens. Fig.~\ref{fig:pipeline}(c) illustrates this process.

\begin{figure}[t]
\small
\centering
    \begin{minipage}{1.02\linewidth}
        \centering
        \includegraphics[width=\textwidth]{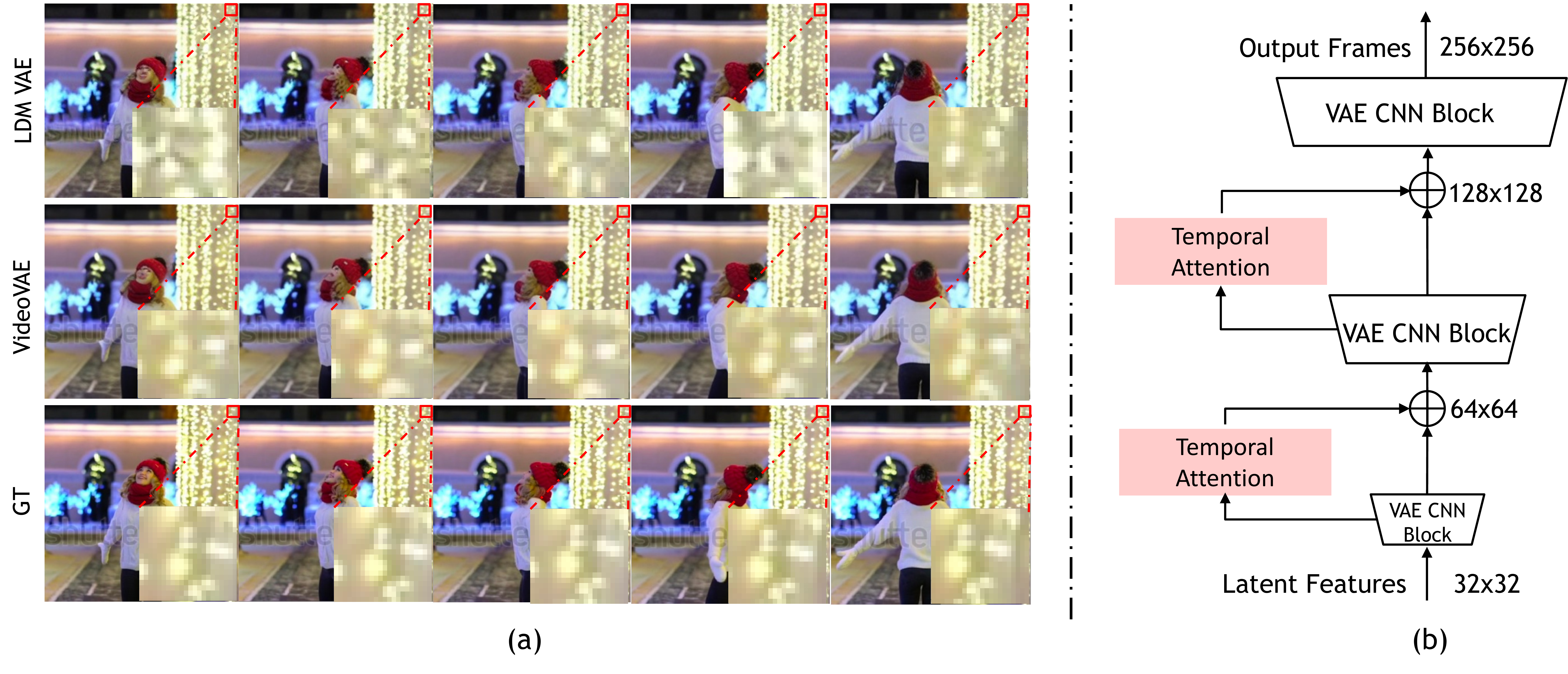}
    \end{minipage}\hfill
\caption{{(a) Our proposed VideoVAE for decoding latent features to video frames can effectively  reducing pixel dithering artifacts, compared with the conventional VAE model. (b) The architecture of our proposed VideoVAE. }
\vspace{-3mm}
}
\label{fig:videovae}
\end{figure}

\subsubsection{Training strategy}
\label{subsubsec:sampling}
\myPara{Frame sampling and training objective.}
%
During training, we first randomly sample a small portion of successive frames (length $L_s$)  from each video and read out its frame-per-second (FPS) metadata. Then, we sample 16 frames uniformly from the selected subset as training data. The length of the selected small portion implicitly indicates the speed of motion changing observed within the sampled 16 frames, \ie the longer the subset is, the faster the scene changes. Thus, we compute $\nu =  \frac{16}{L_s}\cdot \text{FPS}$ as the new FPS of the 16 frames and use it as an input embedding to MagicVideo. Specifically, we  use two linear layers to transform the new FPS $\nu$  into an embedding of dimension $C$:
\begin{equation}
   \text{emb}_{\nu} = \text{Linear}(\text{SiLU}(\text{Linear}(\text{Sin}(\nu)))),
\end{equation}
where $\text{Sin}(\cdot)$ denotes the sinusoidal position embedding~\cite{vaswani2017attention}. $\text{SiLU}$ denotes the sigmoid ReLU \cite{elfwing2018sigmoid}. The embedding $\text{emb}_{\nu} \in \mathbb{R}^{ C}$ will be added to the video frame features, $\vz \in \mathbb{R}^{F\times H \times W \times C}$. 

We directly use the frame-wise reconstruction loss for the model training. Given a sequence of video frames, the loss of a certain sequence is computed as follows, 
\begin{equation}
        \mathcal{L}(\vz_0) = \E_{t}\sum_{i=0}^F {\|\vz_0^i - \epsilon_{\theta}(\vz^i_t, t, \vy)\|_2^2}.
        \label{eq:loss}
\end{equation}

\myPara{Unsupervised training scheme.}
\label{subsec:unsupervised_training}
Text-video pairs are scarce in practice. On the contrary, it is easy to collect abundant high-quality   video-only   data. Motivated by \cite{singer_make--video_2022}, we adopt an unsupervised training strategy where the embeddings of video frames are used as proxies of text conditions to pretrain the model. The embeddings are extracted using the vision encoder of CLIP \cite{clip}. After the unsupervised stage, we finetune the model on a  well-annotated video-text paired dataset. The unsupervised and supervised training use the same training objective as defined in Eqn.~\eqref{eq:loss}.

\subsection{Frame interpolation}
\label{subsec:interpol}

To increase the temporal resolution and make the generated video smoother, we train a separate frame-interpolation network to synthesize unseen frames between two adjacent key frames. The interpolation model is also trained in the latent space under a similar pipeline as the key frame generation. The difference is that the generation   of the interpolated frame features   $\vz$ is conditioned on the adjacent two frames. The conditioning embeddings of the adjacent frames are extracted by CLIP's vision  encoder and injected into the cross-attention layers. 
Besides, we also concatenate the adjacent two frames' latent embeddings to the randomly sampled noise as input to the interpolation model. 
%
We initialize the interpolation U-Net  with the key frame generation model for faster convergence. 
For each pair of two adjacent frames, the interpolation network predicts 3 new intermediate frames between them.

\begin{figure*}[t] 
\small
\centering
    \begin{minipage}{0.52\linewidth}
        \centering
        \includegraphics[width=\textwidth]{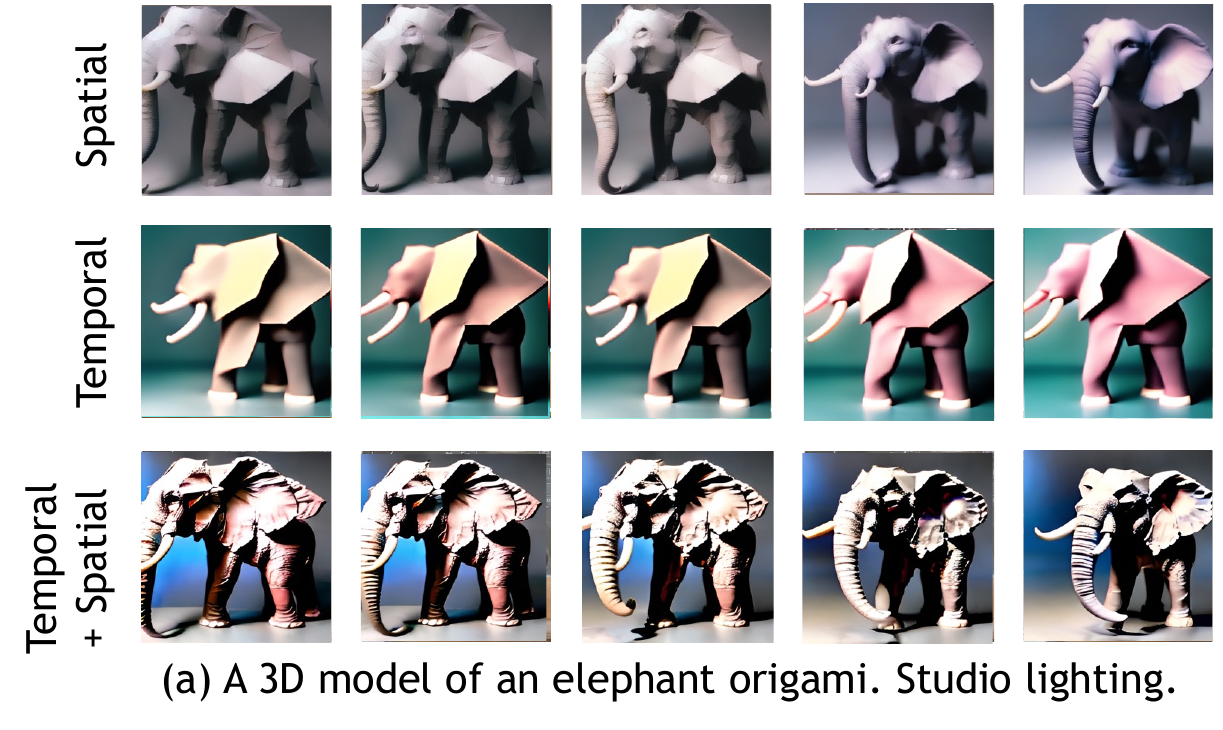}
        
    \end{minipage}\hfill
    \begin{minipage}{0.46\linewidth}
        \centering
        \includegraphics[width=\textwidth]{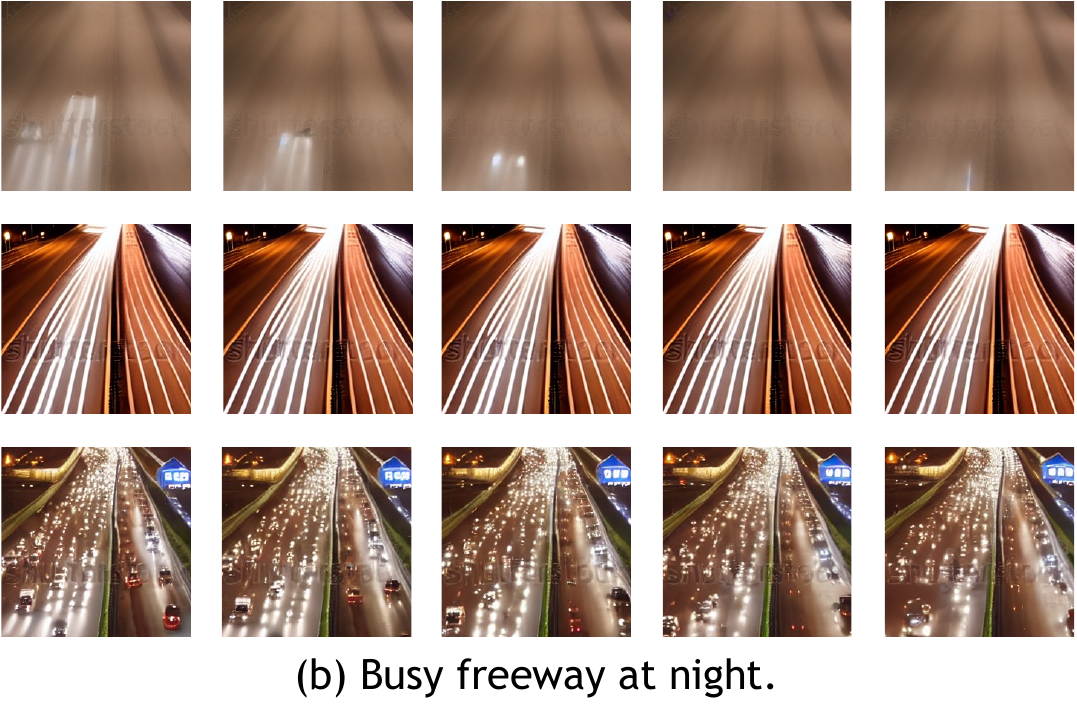}
    \end{minipage}\hfill
    \vspace{-3mm}
\caption{\textbf{Illustration of effects of the attention mechanisms.} Spatial attention helps generate diverse frame contents and temporal attention  tends to guarantee cross-frame consistency. 
}
\vspace{-3mm}
\label{fig:spatiotemporal}
\end{figure*}




\subsection{VideoVAE decoder}
\label{subsec:video_decoder}
In  LDM \cite{rombach_high-resolution_2022}, RGB images are synthesized by decoding latent features via a pre-trained VAE decoder. In practice, we observe pixel dithering   in the generated video frames if we reconstruct videos frame-by-frame via the VAE decoder, leading to visually aesthetic degradation, as shown in Fig.~\ref{fig:videovae}. 

We empirically find the appearance  of dithering   relates to the spatial dimension of the latent features:  using  features with higher   dimension suffers   less     dithering. However, the computational cost will increase if naively increase  the feature spatial dimensions. To improve the visual quality without incurring  much computational overhead, we keep   low dimension of the latent features while adding two   temporal directed attention layers in the decoder   to build a VideoVAE decoder, as shown in Fig.~\ref{fig:videovae}.  We find it effectively alleviates the dithering artifacts. 

\subsection{Super-resolution}
\label{subsec:super_res}
To generate high-resolution videos, we train a diffusion-based super-resolution (SR) model~\cite{saharia2022image} in RGB space to upsample videos from 256$\times$ 256 to 1024$\times$ 1024. The SR model is trained only on image datasets because large-scale high-resolution video datasets are not publicly available and   hard to collect. 
To reduce its computational and memory cost, we train the SR model on 512$\times$512 random crops of 1024$\times$1024 images with 128$\times$128 input frames. During inference, we feed the generated 256$\times$256 frames as the input to generate frames with 1024$\times$1024   dimension. Chitwan et al.~\cite{saharia2022photorealistic} observed noise conditioning augmentation on super-resolution is critical for generating high-fidelity images. Thus, following~\cite{saharia2022photorealistic}, we degrade low-resolution images with  random Gaussian noise   and add the noise level as another conditioning signal of the diffusion model.

\section{Experiments}
\label{sec:exp}

\myPara{Datasets.} We use the weights of LDM~\cite{rombach2022high} pre-trained on Laion 5B \cite{laion5b} to initialize  our video 3D U-net denoising decoder. We then conduct unsupervised training on a subset (10M videos) of HD-VILA-100M \cite{xue2022advancing} and Webvid-10M \cite{bain2021frozen}. We fine-tune the video generation model on a subset of self-collected 7M video-text samples. For the ablation study, we randomly sample a subset of 50k videos from Webvid10M to save the computational cost. When comparing with other methods, we evaluate the zero-shot   performance with text prompt from the test dataset of UCF-101 \cite{soomro2012ucf101}, MSR-VTT \cite{xu2016msr} and calculate the Frechet Inception Distance (FID) \cite{parmar2021cleanfid} and Frechet Video Distance (FVD) \cite{unterthiner2019fvd} with reference to the images in their test dataset. For implementation details, please refer to the sumpplementary material due to space limit.

%

\begin{figure*}[h]
    \centering
    \includegraphics[width=1\textwidth]{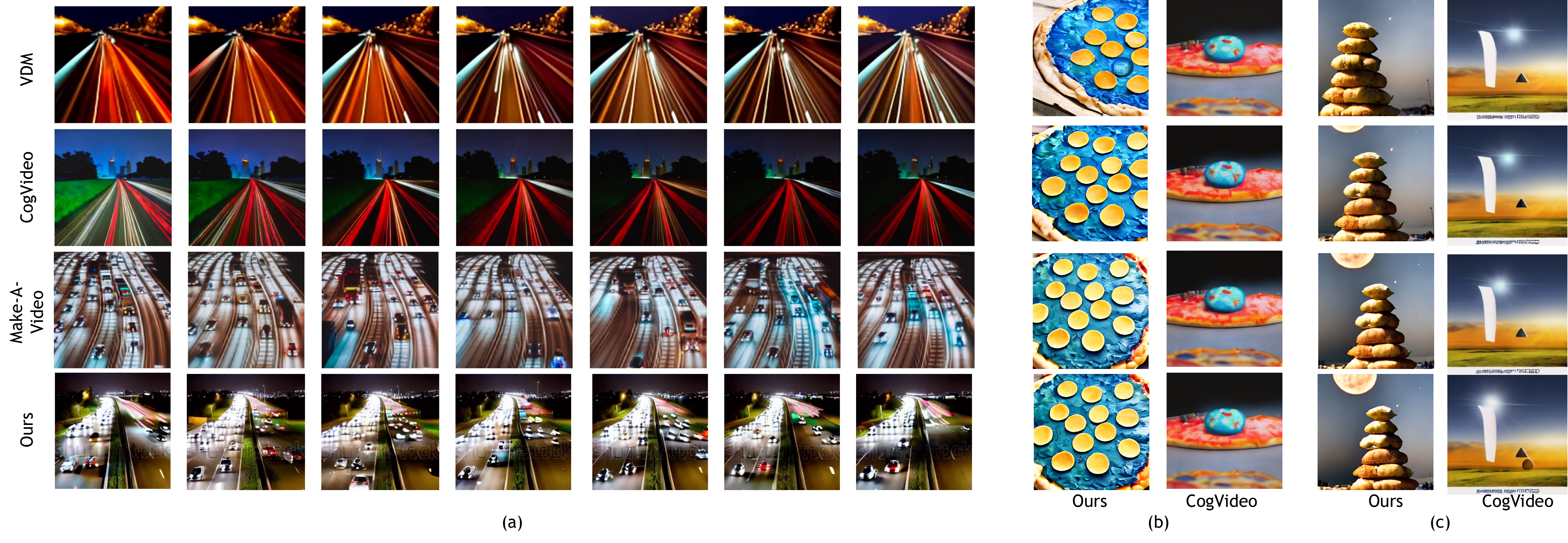}
    \caption{\textbf{Qualitative results comparison with recent strong methods}: VDM \cite{ho_video_2022}, CogVideo \cite{CogVideo}, and Make-A-Video~\cite{singer_make--video_2022}.  (a) The sample videos   are generated with text input ``Busy freeway at night", where the samples of the VDM are taken from~\cite{singer_make--video_2022}; (b) \& (c)  the samples are generated with the text ``A blue colored dog" and ``A pyramid made of falafel with a partial solar eclipse in the background" respectively. See Tab. \ref{tab:eval_human} for detailed comparison with CogVideo. } %
    \label{fig:visual_results}
\end{figure*}
\subsection{Analysis}

\begin{figure}[h] 
\centering
\includegraphics[width=1.0\linewidth]{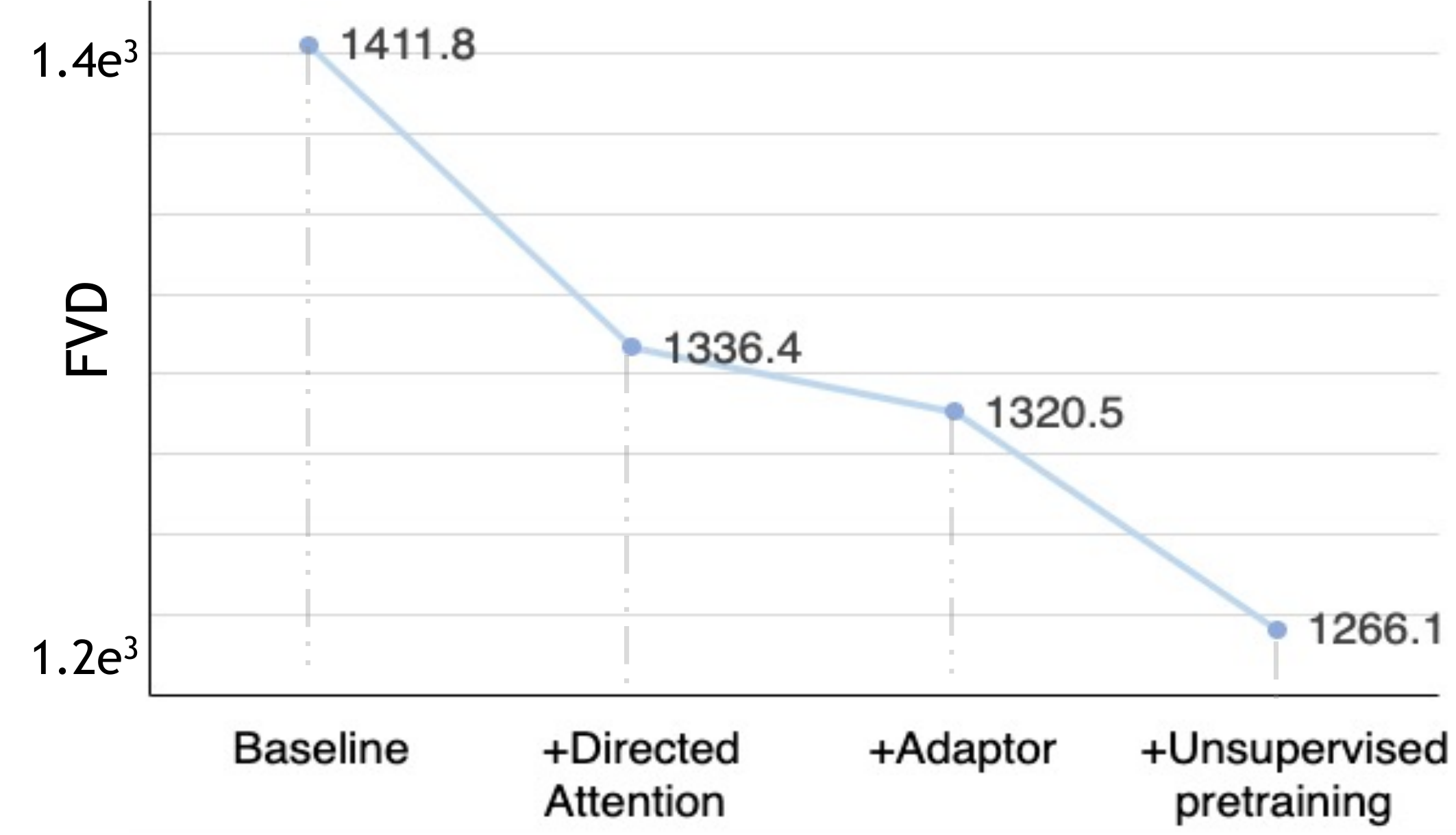}
\caption{Impacts of different components on the model  performance (FVD).}
\label{fig:ablation_FID}
\end{figure}

\myPara{Ablation studies.} We first investigate the impacts of each proposed component via ablation studies. We randomly sample a subset of 1000 video clips from the test dataset of the Webvid-10M dataset and extract 16 frames from each video to form the reference dataset. The results are shown in Fig.~\ref{fig:ablation_FID}. We can observe the directed attention can substantially reduce the FVD. The adaptor not only saves computation cost but also benefits the video generation quality. The unsupervised pre-training can significantly improve the quality\textemdash the FVD is further reduced by around 60.

\begin{figure*}[t!] 
\centering
\includegraphics[width=0.9\linewidth]{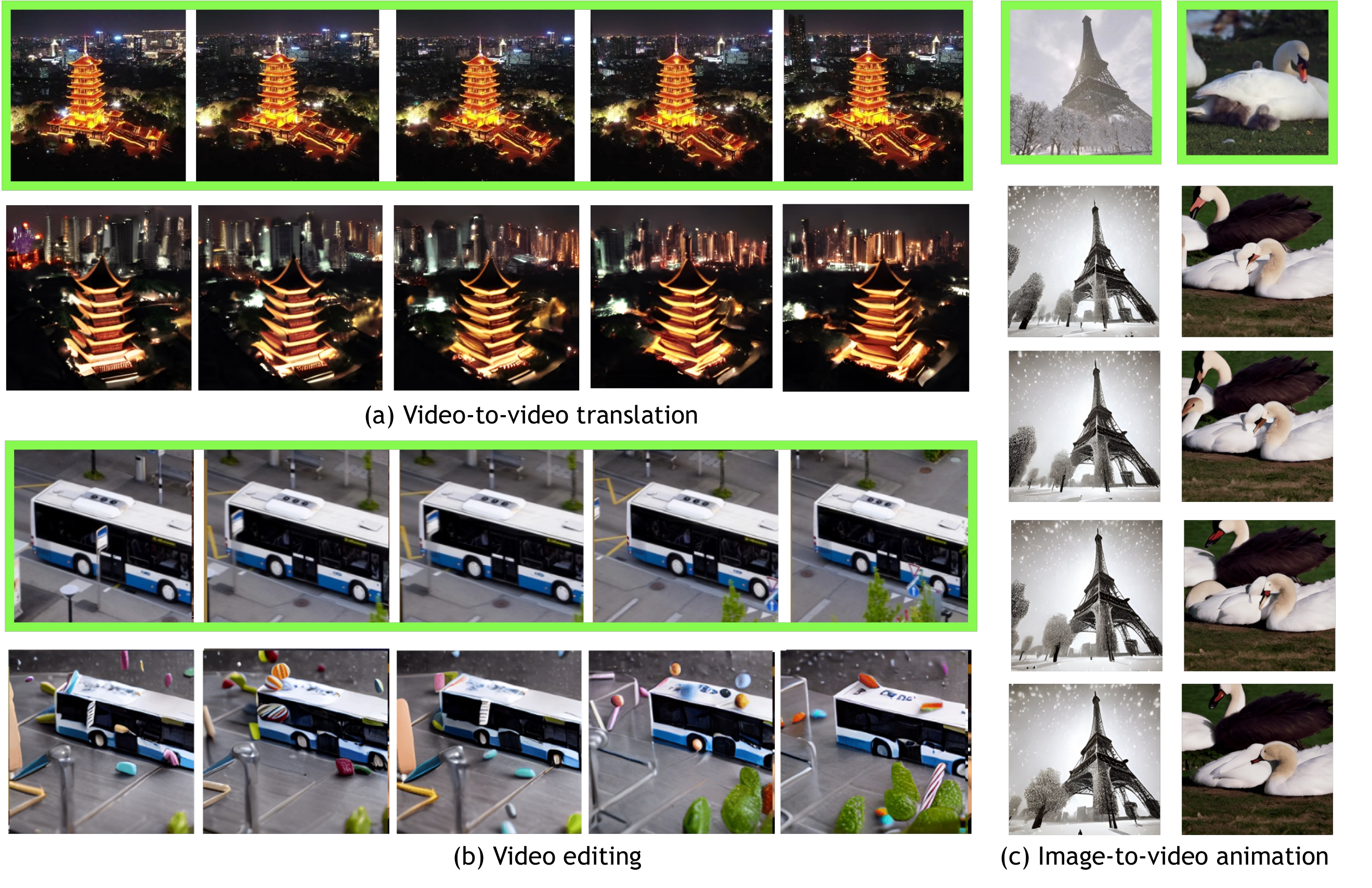}
\vspace{-3mm}
\caption{\textbf{Applications based on MagicVideo.} The {{\bf green  bounding box}} denotes the source image/video frames. (a)    The video variation results. (b)    Given video clip of a driving bus,  we use text prompt of ``candies falling onto the ground" to MagicVideo for video editing. (c) Given an input image, MagicVideo can generate a short relevant video.  }
\vspace{-3mm}
\label{fig:applications}
\end{figure*}

\myPara{Spatial and temporal attention.}
Different from concurrent works \cite{singer_make--video_2022}, we use a dual self-attention design in parallel: one for spatial self-attention and temporal attention learning in parallel. The detailed architecture for the attention block is shown in Fig.~\ref{fig:pipeline}(c). To understand the difference between spatial and temporal attention, we visualize the extracted features from each branch separately, and the results are shown in Fig. \ref{fig:spatiotemporal}. We find that spatial attention helps generate diverse video frames, and temporal attention generates consistent output among the frames. Combining the spatial and temporal attention guarantees the content diversity and consistency for the generated frames.


\subsection{Results}

\myPara{Qualitative evaluation.}
We first evaluate our video generation model on qualitative generation performance and compare it with recent state-of-the-art  models. The visual results are shown in Fig.~\ref{fig:visual_results}, where we compare  with three strong baselines. We want to highlight that \textit{Make-A-Video is a concurrent work}. Compared with CogVideo~\cite{CogVideo} and VDM \cite{ho2022video}, both Make-A-Video and our model can generate videos with richer details. For example, with ``Busy freeway at night" as the text input, the videos generated by CogVideo  and VDM  only show abstract scenes with motion flow without any clear objects (\eg, the cars). Differently,   our MagicVideo can generate complex highway objects such as cars with headlights. Moreover, MagicVideo can even generate the perspective phenomenon\textemdash the video from our model shows clearer vehicles near the camera. \textbf{More samples of the generated videos are provided in the supplementary material}.

\myPara{Quantitative evaluation.}
We also evaluate MagicVideo quantitatively. Specifically, we pre-train  the model   on the Webvid-10M dataset. Then we use the text descriptions of the test data of   MSR-VTT   and the class label of   UCF-101 validation data as the text prompts to generate 16 key frames for each text prompt without fine-tuning. The comparison between MagicVideo and other recent SOTA methods is  shown in   Tab.~\ref{tab:msrvtt} and Tab.~\ref{tab:ucf101}.

\begin{table}[t]
\small
\caption{Video generation evaluation on MSR-VTT. }
\label{tab:msrvtt}
\centering
\begin{tabular}{@{}l|ccc@{}}
Method & Zero-Shot  & FID $\downarrow$ & FVD $\downarrow$  \\
\toprule
NÜWA~\cite{wu2022nuwa} & No &  $47.7$ & $-$ \\
CogVideo~\cite{CogVideo}  & Yes &  $49.0$ & $1294$ \\
MagicVideo (ours) & Yes & $36.5$ & $998$ \\
\end{tabular}
\end{table}

\begin{table}[h]
\small
\caption{Video generation evaluation on UCF-101. }
\label{tab:ucf101}
\centering
\begin{tabular}{@{}l|ccc@{}}
Method & Zero-Shot  & FID $\downarrow$ & FVD $\downarrow$  \\
\toprule
MoCoGAN-HD\cite{tian2021good} & No & - & 700 \\
CogVideo~\cite{CogVideo} (Chinese) & Yes &  $185$ & $751$ \\
CogVideo~\cite{CogVideo} (English) & Yes &  $179 $ & $702$ \\
MagicVideo (ours) & Yes  & $145$ & $655$ \\
\end{tabular}
\end{table}

\begin{table}[t]
\caption{Human evaluation comparison between MaigicVideo (ours) and CogVideo~\cite{CogVideo} on DrawBench \cite{imagen} on 200 prompts. We evaluate the quality of the generated videos based on four aspects: Realism, Faithfulness, Smoothness, and Efficiency (time cost for generating 16 frames). 
%
The numbers in the table show the percentage of samples that the results of our MagicVideo model are better or equal to the samples generated from CogVideo.}
\label{tab:eval_human}
\centering
\footnotesize
\begin{tabular}{cccc}
Realism & Faithfulness & Smoothness & Time cost (mins/16 frames) \\
\toprule
$68$ & $86$ & 96 & 0.3 vs 21\\ 
\end{tabular}
\end{table}


\myPara{Human Evaluation.} We also compare to CogVideo (the state-of-the-art open sourced model) on  DrawBench \cite{imagen} by inviting multiple raters. The results are shown in Tab. \ref{tab:eval_human}.   MagicVideo performs much better than CogVideo \cite{CogVideo} with significantly faster speed.

\subsection{Applications}
We   present three applications based on MagicVideo: i) image-to-video generation: given the an input reference image, generating the videos based on the image;  ii) video variations: generating a  similar video frame sequence based on the input video frames;  and iii) video editing: changing the video frame contents based on the input text prompts. As shown in Fig.~\ref{fig:applications}(a), with a given image input, MagicVideo is able to generate coherent video frames that are closely related to the main context of the single image input. Fig.~\ref{fig:applications}(b) demonstrates that MagicVideo is able to generate variants of a given video input and Fig.~\ref{fig:applications}(c) shows that by adding some text prompt, MagicVideo can be used to edit a given video. More detailed descriptions on the settings of the three applications are put in the supplementary material.

\section{Conclusions}
In this paper, we stepped toward solving the video generation challenge. In particular, we focused on improving the data and computational efficiency of the video generation models. We leveraged the recent latent diffusion model and developed the video generation framework, MagicVideo, in a low-dimensional latent space. Additionally, we introduced several new designs, including the directional attention and the adaptor module, to sufficiently utilize pre-trained image generation models. Finally, we demonstrated MagicVideo indeed generates realistic and smooth videos from a text  description efficiently. 

\myPara{Ethical impact.} Video generation may have significant ethical impacts. Besides the applications on   generative models  for entertainment and art creation, video generation methods are also applicable for malicious purposes by editing videos. However, current deep fake detection technology can detect the fake contents. Another potential issue is using the pre-trained weights from Stable Diffusion \cite{rombach2022high},
which was trained on the LAION dataset \cite{laion5b}. Therefore, it may inherit the LAION dataset   contents with ethical issues  \cite{birhane2021multimodal}.


{
\bibliographystyle{ieee_fullname}
\bibliography{egbib}
}

\end{document}